\documentclass{article}
\usepackage{amsmath}
\usepackage{amsfonts}
\usepackage{amsthm}
\usepackage{amssymb}
\usepackage{bm}
\usepackage{fontawesome5}
\usepackage{pgffor}
\usepackage{multirow}
\usepackage{verbatim}
\usepackage{enumitem}
\usepackage{microtype}
\usepackage{graphicx}
\usepackage{subfigure}
\usepackage{booktabs}
\usepackage{url,hyperref}

\usepackage{mathtools}
\newtheorem{thm}{Theorem}
\newtheorem{lem}[thm]{Lemma}
\newtheorem{prop}[thm]{Proposition}

\newtheorem{defn}[thm]{Definition}
\newtheorem{assu}[thm]{Assumption}

\newtheorem{rem}[thm]{Remark}
\usepackage{algorithm2e}

\usepackage[accepted]{icml2023}

\newcommand\blfootnote[1]{
  \begingroup
  \renewcommand\thefootnote{}\footnote{#1}
  \addtocounter{footnote}{-1}
  \endgroup
}

\newcommand{\ie}{\emph{i.e.}}
\newcommand{\eg}{\emph{e.g.}}

\icmltitlerunning{Nonparametric Iterative Machine Teaching}


\begin{document}
	
	\twocolumn[
	\icmltitle{Nonparametric Iterative Machine Teaching}
	
	\begin{icmlauthorlist}
		\icmlauthor{Chen Zhang}{jlu}
		\icmlauthor{Xiaofeng Cao}{jlu}
		\icmlauthor{Weiyang Liu}{mpi,uc}
		\icmlauthor{Ivor W. Tsang}{ast}
		\icmlauthor{James T. Kwok}{hk}
	\end{icmlauthorlist}
	
	\icmlaffiliation{jlu}{School of Artificial Intelligence, Jilin University, China}
        \icmlaffiliation{mpi}{Max Planck Institute for Intelligent Systems, T\"ubingen, Germany}
        \icmlaffiliation{uc}{University of Cambridge, United Kingdom}
	\icmlaffiliation{ast}{Centre for Frontier AI Research and Institute of High Performance Computing, A*STAR, Singapore}
	\icmlaffiliation{hk}{Hong Kong University of Science and Technology}
	
	\icmlcorrespondingauthor{Xiaofeng Cao}{xiaofengcao@jlu.edu.cn}
	
	\icmlkeywords{Machine Teaching, Iteration}
	
	\vskip 0.3in
	]
	
	 \printAffiliationsAndNotice{}

\begin{abstract}
In this paper, we consider the problem of Iterative Machine Teaching (IMT), where the teacher provides examples to the learner iteratively such that the learner can achieve fast convergence to a target model. However, existing IMT algorithms are solely based on parameterized families of target models. They mainly focus on convergence in the parameter space, resulting in difficulty when the target models are defined to be functions without dependency on parameters. To address such a limitation, we study a more general task -- Nonparametric Iterative Machine Teaching (NIMT), which aims to teach nonparametric target models to learners in an iterative fashion. Unlike parametric IMT that merely operates in the parameter space, we cast NIMT as a functional optimization problem in the function space. To solve it, we propose both random and greedy functional teaching algorithms. We obtain the iterative teaching dimension (ITD) of the random teaching algorithm under proper assumptions, which serves as a uniform upper bound of ITD in NIMT. Further, the greedy teaching algorithm has a significantly lower ITD, which reaches a tighter upper bound of ITD in NIMT. Finally, we verify the correctness of our theoretical findings with extensive experiments in nonparametric scenarios.
\end{abstract}

\section{Introduction} \label{introduction}

Machine teaching (MT)~\cite{zhu2015machine, zhu2018overview} is the study of how to design the optimal teaching set, typically with minimal examples, so that learners can quickly learn target models based on these examples. It can be considered an inverse problem of machine learning, where machine learning aims to learn model parameters from a dataset, while MT aims to find a minimal dataset from the target model parameters. MT has proven to be useful in various domains, including robustness~\cite{alfeld2016data, alfeld2017explicit, ma2019policy, rakhsha2020policy}, crowd sourcing~\cite{singla2013actively, singla2014near, zhou2018unlearn, zhou2020crowd,Collins2023HILLM}, and computer vision~\cite{wang2021gradient,wang2021machine}.

Considering the interaction manner between teachers and learners, MT can be conducted in either batch \cite{zhu2013machine, zhu2015machine, liu2016teaching, mansouri2019preference} or iterative \cite{liu2017iterative, liu2018towards} fashion. Batch MT only allows the teacher to interact with the learner once. The teacher constructs a teaching dataset and feeds it to the learner in one shot. The learner will learn a target model from this dataset. The minimal number of examples in this teaching set is called \textit{teaching dimension} \cite{goldman1995complexity}. In contrast, an iterative teacher would feed examples sequentially based on current status of the iterative learner, which further takes the optimization algorithm into consideration. The number of iterations, \ie, the length of this teaching sequence is defined as \textit{iterative teaching dimension} (ITD) \cite{liu2017iterative, liu2018towards}.\blfootnote{Our source code is available at \url{https://github.com/chen2hang/NonparametricTeaching}.}

\begin{figure}[t] 
	\vskip -12pt
	\subfigbottomskip=-6.5pt
	\centering
	\subfigure[Parametric IMT]{\includegraphics[width=0.48\linewidth]{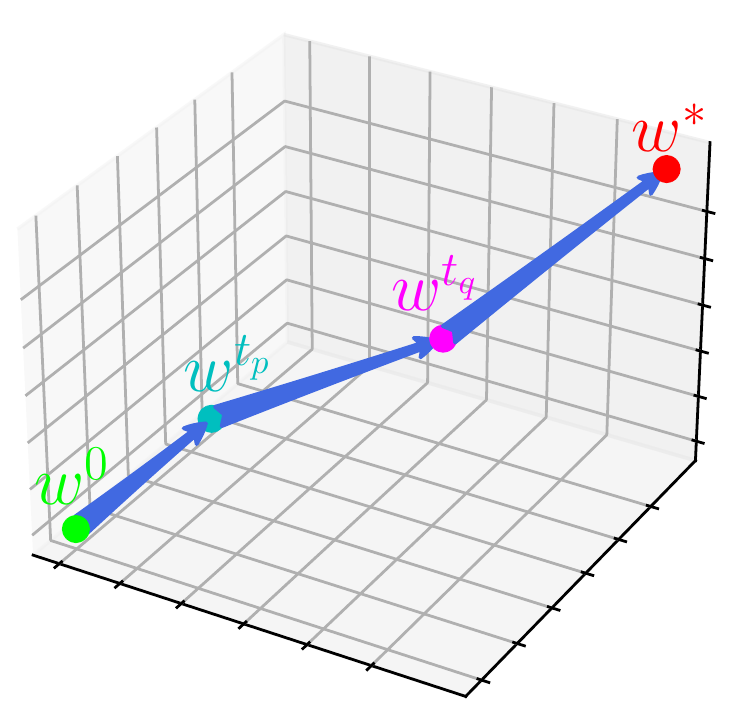}}
	\subfigure[Nonparametric IMT]{\includegraphics[width=0.48\linewidth]{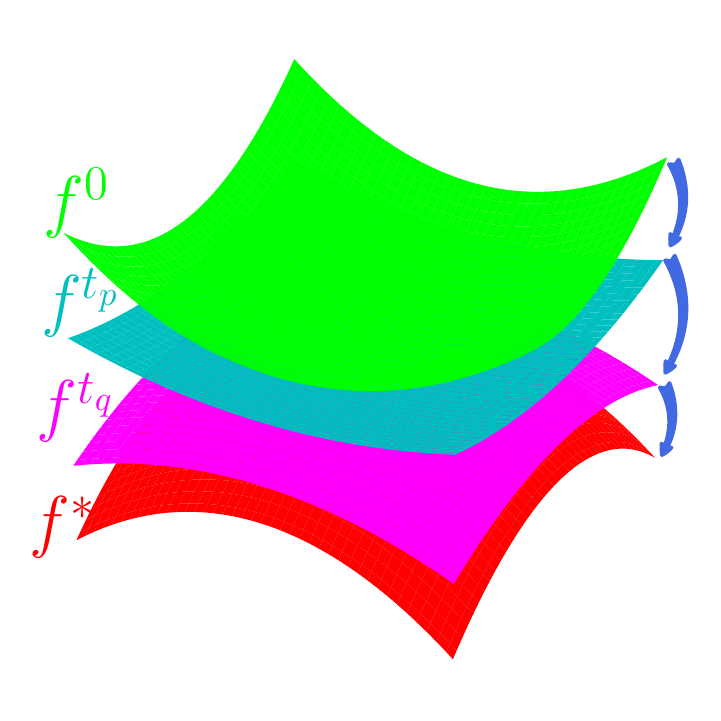}}
	\vskip 2pt
	\caption{Comparison between parametric and nonparametric IMT in 3D space. (a): Parameters are precisely vectors represented by a point in 3D space, which would be updated gradually towards $w^*$. (b): Nonparametric model $f$ can be denoted by a surface in 3D, which would evolve in more complicated fashion.}
	\label{paravsnon}
	\vskip -0.1in
\end{figure}

The majority of current research on iterative machine teaching (IMT) \cite{liu2017iterative, liu2018towards, xu2021locality, wang2021gradient} focuses on the convergence to target models (\ie, functions) $f$ which are usually parameterized by a set of parameters $\bm{w}$ as it assumes that $f$ can be represented by $\bm{w}$, \eg, $f(\bm{x})=\langle \bm{w}, \bm{x}\rangle$ with input $\bm{x}$. However, there may exist cases where the mapping from input to output cannot be parameterized in terms of $\bm{w}$, for example, $f$ is defined in a nonparametric fashion~\cite{hollander2013nonparametric, corder2014nonparametric, zhu2018overview}. Especially in general and more realistic problems (\eg, \cite{genevay2016stochastic, blei2017variational, dvurechenskii2018decentralize}), the assumption of parametric learners may not hold. Here comes a natural question: \textit{Can the teacher efficiently guide iterative learners to parameter-free target models?} Our answer is \textit{Yes}. We seek to guide iterative learners to achieve fast convergence to a  nonparametric target function $f^*$. Figure~\ref{paravsnon} provides an intuitive comparison between parametric and nonparametric iterative teaching in a 3-dimensional space.

Shifting our focus to functions, we formulate NIMT as an instance of functional optimization problem~\cite{singer1974theory, zoppoli2002approximating, mroueh2019sobolev, shen2020sinkhorn}, and then derive two algorithms (one picks examples randomly, and the other picks examples in an greedy fashion). Without loss of generality, we are mainly concerned with the Reproducing Kernel Hilbert Space (RKHS) in this paper. We start with a simple baseline algorithm, called \textbf{R}andom \textbf{F}unctional \textbf{T}eaching (RFT), which essentially adopts uniform sampling and serves as a functional analogue of stochastic gradient descent \cite{ruder2016overview, hardt2016train}.  In the context of IMT, we analyze the functional gradient descent method~\cite{mason1999boosting, shen2020sinkhorn} in RKHS, and then find that based on the chain rule for functional gradients \cite{ gelfand2000calculus, coleman2012calculus}, the gradient in NIMT can be expressed by the multiplication of a scalar governing the magnitude and the kernel function with the teaching example as its argument. Therefore, steepening gradients is equivalent to maximizing that scalar, which naturally leads to our greedy algorithm -- \textbf{G}reedy \textbf{FT} (GFT). GFT picks examples evaluated at the point where the target and current models reach their maximal difference \cite{arbel2019maximum, cormen2022introduction}. Furthermore, under mild assumptions, we theoretically prove the convergence of both RFT and GFT, and then show that the ITD of GFT is lower than that of RFT. This concludes that GFT yields a tighter upper bound for ITD. Finally, we validate our theoretical findings with a number of experiments in both synthetic and real-world datasets under nonparametric scenarios. To summarize, the contributions of our work are listed as follows.
\begin{itemize}[leftmargin=*,nosep]
\setlength\itemsep{0.56em}
\item To our knowledge, we are the first to comprehensively study Nonparametric Iterative Machine Teaching (NIMT), which focuses on exploring iterative algorithms for teaching parameter-free target models from the optimization perspective. Instead of operating in the finite-dimensional space of parameters, we formulate NIMT as a functional optimization in the space of infinite-dimensional functions, a more general space of models (\ie, RKHS is considered), in Section~\ref{ts}. NIMT is a natural generation of IMT~\cite{liu2017iterative}, shifting the parametric paradigm to a nonparametric one.
\item We propose two teaching algorithms (RFT and GFT). RFT is based on random sampling with ground truth labels, and the derivation of GFT is based on the maximization of the informative scalar introduced in Proposition~\ref{gs} in order to steepen gradients. These two teaching algorithms proposed in Section~\ref{ta} fill the gap for teaching nonparametric learners in IMT.
\item We theoretically analyze the asymptotic behavior of both RFT and GFT in Section~\ref{ftca}. We prove that per-iteration reduction of loss $\mathcal{L}$ for RFT and GFT has a negative upper bound expressed by the discrepancy of iterative teaching defined in Definition~\ref{disc}, and we derive that the ITD of GFT is $\mathcal{O}(\psi(\frac{2\mathcal{L}(f^0)}{\tilde{\eta}\epsilon}))$ (detailed notations are introduced in the subsequent sections), which is shown to be lower than the ITD of RFT, $\mathcal{O}(2\mathcal{L}(f^0)/\left(\tilde{\eta}\epsilon\right))$.
\end{itemize}

\vspace{0.5mm}

\section{Related Work}

\textbf{Machine teaching}. There has been a recent growth of interest in the research of machine teaching \cite{zhu2015machine, zhu2018overview, liu2017iterative, liu2018towards, wang2021gradient}. Batch machine teaching studies behaviors of version space learners \cite{chen2018understanding, tabibian2019enhancing}, linear learners \cite{liu2016teaching}, reinforcement learners \cite{kamalaruban2019interactive, zhang2020sample} along with forgetful learners \cite{hunziker2018teaching, liu2018towards} and multiple learners \cite{zhu2017no}. Further, taking the learner's optimization algorithm into consideration, iterative teaching has been recently studied~\cite{liu2017iterative,liu2018towards,peltola2019machine, lessard2019optimal,Liu2021LAST,xu2021locality,qiu2022iterative}. \cite{Liu2021LAST} considers a label synthesis teacher and \cite{qiu2022iterative} proposes a generative teacher. \cite{xu2021locality} improves the scalability and efficiency of the iterative teaching algorithm with locality-sensitive sampling.  Different from existing works that focus on parametric learners, we aim to teach a nonparametric learner. In this regime, One of the most related work is \cite{mansouri2019preference} which analyzes sequential teaching from the perspective of hypothesis pruning without specifying a parameter for hypothesis. In contrast, this work systematically investigates nonparametric teaching from the optimization perspective. Besides, \cite{kumar2021teaching, qian2022teaching} are also highly related, since they study non-gradient-based kernel learners under the batch setting. However, they are not strictly nonparametric teaching since they assume the hypothesis is determined by parameters, and they cannot produce an iterative algorithm for teaching parameter-free mappings. In contrast, we study a more general task -- nonparametric iterative machine teaching, and propose practical iterative functional teaching algorithms.

\textbf{Functional optimization}. Allowing non-parametrically defined mapping from input to output, functional optimization \cite{singer1974theory, becke1988density, singer1974theory, friedman2001greedy, zoppoli2002approximating, smanski2014functional, zhang2020black} over more general space of functions, including RKHS, Sobolev space \cite{adams2003sobolev} and Fréchet space \cite{narici2010topological}, is a foundational and meaningful task across many domains, such as barycenter problem \cite{shen2020sinkhorn, ye2017fast}, variational inference \cite{liu2016stein, liu2017stein} and GAN training \cite{mroueh2019sobolev}. \cite{nitanda2018functional, nitanda2020functional} make an interesting connection between functional gradient boosting and residual networks~\cite{he2016deep}. We observe that the functional gradient descent algorithm \cite{mason1999boosting, mason1999functional, coleman2012calculus} for functional optimization in RKHS is well studied because of some regular properties. The iterative interaction \cite{liu2017iterative, liu2018towards} between teachers and learners exhibits intriguing similarities to the functional gradient descent algorithm in terms of its gradual improvement. Inspired by such similarities, NIMT starts by analyzing functional gradient descent and then designs algorithms for choosing optimal teaching examples under the iterative teaching framework \cite{liu2017iterative, liu2018towards,Liu2021LAST,qiu2022iterative}.

\section{Notations}
Let $\mathcal{X}\subseteq\mathbb{R}^n$ be a $n$ dimensional feature space and $\mathcal{Y}\subseteq\mathbb{R}(\text{Regression})\,\text{or}\,\mathcal{Y}=\left\{-1,1\right\}(\text{Classification})$ be a label space. A teaching example refers to a pair of data (\eg, image) and label $(\bm{x},y)\in\mathcal{X}\times\mathcal{Y}$. A length-$T$ teaching sequence is defined as $\mathcal{D}=\{(\bm{x}^1,y^1),\dots(\bm{x}^T,y^T)\}=\{(\bm{x}^i,y^i)\}_{i=1}^T$. The collection of potential teaching sequences is denoted by $\mathbb{D}$ which includes all teaching sequences, \ie, $\mathcal{D}\in \mathbb{D}$ and is also called the knowledge domain of teachers \cite{liu2017iterative,liu2018towards}. 

This paper considers a specific function space -- the Reproducing Kernel Hilbert Space, and therefore models are assumed to be mappings in RKHS $f\in \mathcal{H}:\mathcal{X}\mapsto\mathcal{Y}$. This assumption is widely adopted in general functional optimization, \eg, \cite{liu2016stein, mroueh2019sobolev, arbel2019maximum, shen2020sinkhorn}. Operating under RKHS where point evaluation is a continuous linear functional allows us to quantify the iteration quality, which is crucial for convergence analysis. Given a target model\footnote{We assume that both $f^0$ and $f^*$ are from the same RHKS such that $f^*$ is realizable. Generally, $f^*$ can be assigned arbitrarily, but for the convergence to the target model, we consider the projection of $f^*$ into the RHKS constructed by $\mathcal{X}$ with specific kernels.} $f^*\in \mathcal{H}$, one can uniquely represent a teaching example $(\bm{x}^\dagger,y^\dagger) $ by its feature $\bm{x}^\dagger$ for brevity since its label is precisely $y^\dagger=f^*(\bm{x}^\dagger)$.

Let $K(\bm{x},\bm{x}'): \mathcal{X}\times\mathcal{X}\mapsto\mathbb{R}$ be a positive definite kernel function. Equivalently, $K(\bm{x},\bm{x}')=K_{\bm{x}}(\bm{x}')=K_{\bm{x}'}(\bm{x})$ and $K_{\bm{x}}(\cdot)$ can be abbreviated as $K_{\bm{x}}$. The RKHS $\mathcal{H}$ determined by $K(\bm{x},\bm{x}')$ is the closure of linear span $\{f:f(\cdot)=\sum_{i=1}^{r}\alpha_i K(\bm{x}_i,\cdot),\alpha_i\in\mathbb{R},r\in\mathbb{N},\bm{x}_i\in\mathcal{X}\}$ equipped with inner product $\langle f,g\rangle_\mathcal{H}=\sum_{ij}\alpha_i\beta_j K(\bm{x}_i,\bm{x}_j)$ when $g=\sum_{j}\beta_j K_{\bm{x}_j}$. NIMT reduces to parameterized IMT if we use a linear kernel:  $K(\bm{x},\bm{x}') = \langle \bm{x},\bm{x}'\rangle + 1$ \cite{hofmann2008kernel}. With the Riesz–Fréchet representation theorem \cite{lax2002functional, scholkopf2002learning}, the evaluation functional is defined as follows:
\begin{defn}
	\label{efl}
	For a reproducing kernel Hilbert space $\mathcal{H}$ with a positive definite kernel $K_{\bm{x}}\in\mathcal{H}$, we define the evaluation functional $E_{\bm{x}}[\cdot]:\mathcal{H}\mapsto\mathbb{R}$ as
	\begin{equation}
		E_{\bm{x}}[f]=\langle f, K_{\bm{x}}(\cdot)\rangle_\mathcal{H}=f(\bm{x}),f\in\mathcal{H}.
	\end{equation}
\end{defn}
Additionally, for a functional $F:\mathcal{H}\mapsto\mathbb{R}$, the Fréchet derivative \cite{coleman2012calculus, liu2017stein, shen2020sinkhorn} of $F$ is given as follows:
\begin{defn} (Fréchet derivative in RKHS)
	\label{defn1}
	For a functional $F:\mathcal{H}\mapsto\mathbb{R}$, its Fréchet derivative $\nabla_f F[f]$ at $f\in\mathcal{H}$ is defined implicitly as $F[f+\epsilon g]=F[f]+\epsilon\langle\nabla_f F[f],g\rangle_\mathcal{H}+\mathcal{O}(\epsilon^2)$ for any $g\in\mathcal{H}$ and $\epsilon\in\mathbb{R}$, which is a function in $\mathcal{H}$.
\end{defn}

\section{Nonparametric Iterative Machine Teaching}

We start by formulating NIMT as a nested functional minimization (Eq.~\ref{eq1}). Then we present a natural baseline called random functional teaching, which samples data randomly (Algorithm~\ref{aaft}). After gaining an insight from functional gradient (Proposition~\ref{gs}), we propose the greedy teaching algorithm, called greedy functional teaching, which searches examples with steeper gradients (Algorithm~\ref{aaft}). Finally, we analyze the ITD for both RFT and GFT.

\subsection{Teaching settings}\label{ts}

Different from the parametric cases \cite{liu2017iterative, zhu2018overview} reviewed in Appendix~\ref{parat}, we define NIMT as a functional minimization over $\mathbb{D}$ in RKHS:
\begin{equation}\label{eq1}
	\begin{aligned}
 \mathcal{D}^*=\underset{\mathcal{D}\in\mathbb{D}}{\arg\min}&\quad \mathcal{M}(\hat{f},f^*)+\lambda\cdot \text{len}(\mathcal{D})\\
		\text{s.t.}\quad&\hat{f}=\mathcal{A}(\mathcal{D})
	\end{aligned},
\end{equation}
where $\mathcal{M}$ denotes a discrepancy measure, $\text{len}(\mathcal{D})$, 
which is regularized by a constant $\lambda$, is the length of the teaching sequence $\mathcal{D}$, and $\mathcal{A}$ represents the learning algorithm of learners. In fact, $\text{len}(\mathcal{D})$ essentially is the count of iterations, \ie, the ITD defined in \cite{liu2017iterative}. Specifically, we are concerned with $L_2$ norm defined in RKHS as the discrepancy measure $\mathcal{M}(\hat{f},f^*)=\|\hat{f}-f^*\|_\mathcal{H}$, and empirical risk minimization as the learning algorithm $\mathcal{A}(\mathcal{D})$ as follows:
\begin{equation}
	\label{la}
	\hat{f^*}=\underset {f\in\mathcal{H}}{\arg\min}\,\mathbb{E}_{(\bm{x},y)}\left\{\mathcal{L}(f(\bm{x}),y)\right\},
\end{equation}
where we have the joint sampling distribution $(\bm{x},y)\sim\mathbb{P}(\bm{x},y)$ and the convex loss function $\mathcal{L}$. It is optimized by functional gradient descent:
\begin{equation}
	\label{opta}
	f^{t+1}\gets f^t-\eta^t\mathcal{G}(\mathcal{L};f^t;(\bm{x}^t,y^t)),
\end{equation}
where $t=0,1,\dots,\text{len}(\mathcal{D})$ is the iteration index, $\eta^t>0$ is the learning rate at $t$-th iteration (a small constant) and $\mathcal{G}$ denotes the gradient functional evaluated at $(\bm{x}^t,y^t)$.

Compared to the white-box setting where teachers know all information about learners \cite{liu2017iterative, Liu2021LAST, xu2021locality}, this paper considers a more practical gray-box teaching setting, where teachers have no access to the learning rate $\eta$, specific loss function $\mathcal{L}$ but are able to track $f^t$. For interaction, we only allow teachers to communicate with learners via teaching examples in $\mathcal{D}$. For teachers with different knowledge domains, we start by deriving the theoretical findings for synthesis-based teachers~\cite{liu2017iterative}, and then extend them to the most practical pool-based teachers discussed in Remark~\ref{pbt}. Finally, we study the empirical performance of our method.

\subsection{Functional teaching algorithms}\label{ta}

\textbf{Random Functional Teaching.} It is straightforward for teachers to pick examples randomly and feed them to learners, which derives a simple teaching baseline called Random Functional Teaching. Given a nonparametric target model $f^*$, RFT algorithm is to give learners $\mathcal{D}=\{(\bm{x}^i,y^i)\}_{i=1}^{\text{ITD}_\text{RFT}}$ where $\bm{x}^i\in\mathcal{X}$ is picked randomly, $y^i = f^*(\bm{x}^i)$ and $\text{ITD}_\text{RFT}$ denotes the ITD of RFT. RFT forms a functional counterpart of SGD \cite{ruder2016overview, hardt2016train}, and RFT provides ground truth $y^i = f^*(\bm{x}^i)$ as $f^*$ is known. Therefore, it is natural to consider RFT as a very fundamental baseline when comparing against other functional teaching algorithms. Pseudo code is in Algorithm~\ref{aaft}. 

\textbf{Greedy Functional Teaching.} With Fréchet derivative in RKHS (Definition~\ref{defn1}), we introduce Chain Rule for functional gradients \cite{ gelfand2000calculus} as a Lemma.
\begin{lem}(Chain rule for functional gradients)
	\label{cr}
	For differentiable functions $G: \mathbb{R}\mapsto\mathbb{R}$ that are functions of functionals $F$, $G(F[f])$, the expression 
	\begin{equation}
		\nabla_f G(F[f])=\frac{\partial G(F[f])}{\partial F[f]}\cdot 	\nabla_f F[f]
	\end{equation}
	is usually referred to as the chain rule.
\end{lem}

For derivative of evaluation functional \cite{coleman2012calculus}, we provide Lemma~\ref{ef} whose proof is deferred to Appendix~\ref{pef}.
\begin{lem}
	\label{ef}
	For an evaluation functional $E_{\bm{x}}[f]=f(\bm{x}):\mathcal{H}\mapsto\mathbb{R}$, its gradient is $\nabla_f E_{\bm{x}}[f] = K_{\bm{x}}$.
\end{lem}

$f$ can be viewed as the argument and the loss function $\mathcal{L}$ of interest in NIMT is precisely a functional. Consequently, with Lemma~\ref{cr} and \ref{ef}, we gain a critical insight of functional gradients of $\mathcal{L}$ \cite{mason1999boosting, coleman2012calculus}.
\begin{prop}
	\label{gs}
	 Given a certain example $(\bm{x},y)$, the gradient $\mathcal{G}$ of loss function $\mathcal{L}$ w.r.t. the model $f$ can be expressed as a scalar times a unit kernel:
	 \begin{eqnarray}\label{gexp}
		\mathcal{G}(\mathcal{L};f;(\bm{x},y))=\left.\frac{\partial\mathcal{L}}{\partial f}\right|_{f(\bm{x}),y} \|K _{\bm{x}}\|_\mathcal{H} \cdot \frac{K_{\bm{x}}}{\|K_{\bm{x}}\|_\mathcal{H}}.
	\end{eqnarray}
\end{prop}

Proposition~\ref{gs} suggests that the functional gradient is fundamentally determined by an informative real number $\left.\partial\mathcal{L}/\partial f\right|_{f(\bm{x}),y}\|K_{\bm{x}}\|_\mathcal{H}$ controlling the magnitude of $\mathcal{G}$ and a unit kernel $K_{\bm{x}}/\|K_{\bm{x}}\|_\mathcal{H}$ governing the direction \cite{coleman2012calculus}. For ease of understanding, such a unit kernel can be viewed as a unit vector in infinite dimensional space (a counterpart of a unit vector in the Euclidean space) since a model can be represented by a infinite series of functions in RKHS, $f=\sum_{i}^{\infty}\alpha_i K_{\bm{x}_i}$ \cite{steinwart2008support}. 

In IMT \cite{liu2017iterative}, the target is to achieve fast convergence (maximal reduction of iteration number) by designing the optimal iterative algorithms for example selection. It is natural to consider the properties of the optimal example for reducing ITD at each iteration. This is answered by Theorem~\ref{optthm} proved in Appendix~\ref{poptthm}.

\begin{thm}\label{optthm}
	Given a nonparametric target model $f^*$, let $(\bm{x}^t,y^t)$ be a fed example at $t$-th iteration and $({\bm{x}^t}^*,{y^t}^*)$ be the optimal one with the steepest gradient towards $f^*$:
	\begin{eqnarray} \label{optx}
		&&\left({\bm{x}^t}^*,{y^t}^*\right) = \underset {\bm{x}^t\in\mathcal{X},y^t\in\mathcal{Y}}{\arg\min}\nonumber\\
		&&\qquad\left\|f^t-	\eta^t \mathcal{G}(\mathcal{L};f^t;(\bm{x}^t,y^t))-f^*\right\|^2_\mathcal{H}.
	\end{eqnarray}
	We denote $\mathcal{G}^t\coloneqq\mathcal{G}(\mathcal{L};f^t;(\bm{x}^t,y^t))$ and ${\mathcal{G}^t}^*\coloneqq\mathcal{G}(\mathcal{L};f^t;({\bm{x}^t}^*,{y^t}^*))$, and then the following holds
	\begin{eqnarray}\label{optd}
		\langle {\mathcal{G}^t}^*-{\mathcal{G}^t},f^t-f^*\rangle_\mathcal{H}\geq0.
	\end{eqnarray}
\end{thm}

Eq.~\ref{optd} indicates a property of ${\mathcal{G}^t}^*$ corresponding to ${\bm{x}^t}^*$, which is independent of explicit $\eta$ and specific $\mathcal{L}$ and adapts to gray-box learners. Besides, Theorem~\ref{optthm} intuitively tells that ${\mathcal{G}^t}^*-{\mathcal{G}^t}$ and $f^t-f^*$ share the same direction. That means if $f^t \geq f^*$, the example with the largest gradient ${\mathcal{G}^t}^*\geq\mathcal{G}^t\geq0$ would be selected as the optimal example to minimize Eq.~\ref{optx}. For the case of $f^t\leq f^*$, the gradient of the optimal example should be the smallest one, \ie, ${\mathcal{G}^t}^*\leq\mathcal{G}^t\leq0$. In a nutshell, the gradient norm at the optimal example should be maximal at every iteration.

Combining Proposition~\ref{gs} and results in Theorem~\ref{optthm}, maximizing gradient norm written in Eq.~\ref{gexp} derives our greedy functional teaching algorithm, namely Greedy-1 Functional Teaching (GFT-1):

Given a nonparametric target model $f^*$, GFT-1 is to pick the example satisfying 
	\begin{eqnarray}\label{eqaft1}
		\resizebox{.9\hsize}{!}{$\left({\bm{x}^t}^*=\underset{\bm{x}^t\in\mathcal{X}}{\arg\max}\left\|\left.\frac{\partial\mathcal{L}}{\partial f}\right|_{f^t(\bm{x}^t),y^t=f^*(\bm{x}^t)}K \left(\bm{x}^t,\cdot\right)\right\|_\mathcal{H},{y^t}^*=f^*\left({\bm{x}^t}^*\right)\right)$}
	\end{eqnarray}
	as the optimal one to learners at $t$-th iteration, and $t=0,1,\dots,\text{ITD}_\text{GFT}$ where  $\text{ITD}_\text{GFT}$ is the ITD of GFT.

Practically, we can simplify it as 
\begin{equation}\label{aft1prac}
	\resizebox{.85\hsize}{!}{$\left({\bm{x}^t}^*=\underset{\bm{x}^t\in\mathcal{X}}{\arg\max}\left|\left.\frac{\partial\mathcal{L}}{\partial f}\right|_{f^t(\bm{x}^t),y^t=f^*(\bm{x}^t)}\right|,{y^t}^*=f^*\left({\bm{x}^t}^*\right)\right)$}
\end{equation}
to save computational cost when choosing normalized kernel functions $\|K_{\bm{x}}\|_\mathcal{H}\approx1$ or ignoring the trivial influence from $\|K_{\bm{x}}\|_\mathcal{H}$ when the values of $\|K_{\bm{x}}\|_\mathcal{H}$ are the same for all $\bm{x}\in\mathcal{X}$. Since $\partial\mathcal{L}/\partial f$ has positive correlation with $\|f-f^*\|_\mathcal{H}$: $\partial\mathcal{L}/\partial f$ decrease as $f$ gradually approaches $f^*$ \cite{boyd2004convex, coleman2012calculus}, it is computationally plausible to maximize $|f(\bm{x})-f^*(\bm{x})|$ rather than $\left. \partial\mathcal{L}/\partial f\right|_{f^t(\bm{x}^t),y^t}$ directly, such that GFT-1 also can be implemented under the gray-box setting where $\mathcal{L}$ and $\eta$ could be unknown. Maximizing $|f(\bm{x})-f^*(\bm{x})|$ is easy to compute, since it avoids calculation of the partial derivative when example selection. Compared to RFT, GFT selects examples with a greedy strategy for fast convergence.

\begin{algorithm}[tb]
	\caption{Random / Greedy Functional Teaching}
	\label{aaft}
		 {\bfseries Input:} Target $f^*$, initial $f^0$, per-iteration pack size $k$, small constant $\epsilon>0$ and maximal iteration number $T$.
		 \BlankLine
		Set $f^t\gets f^0$, $t=0$.
		\BlankLine
		\While{$t\leq T$ {\rm and} $\|f^t-f^*\|_\mathcal{H}\geq\epsilon$}{
		\BlankLine
		 \textbf{The teacher} selects $k$ teaching examples:
		 
		 Initialize the pack of teaching examples $\mathcal{K}=\emptyset$;

		\For{\textup{$j=1$ {\bfseries to} $k$}}{
		 (\textbf{RFT}) 1. Pick ${\bm{x}^t_j}^*\in\mathcal{X}$ randomly;
		 \BlankLine
		 (\textbf{GFT}) 1. Pick ${\bm{x}^t_j}^*$ with the maximal difference between $f^t$ and $f^*$: $${\bm{x}^t_j}^*=\underset{\bm{x}^t_i\in\mathcal{X}-\{{\bm{x}^t_i}^*\}_{i=1}^{j-1}}{\arg\max}\left|f^t(\bm{x}^t_i)-f^*(\bm{x}^t_i)\right|;$$\\
		2. Add $\left({\bm{x}^t_j}^*,{y^t_j}^*=f^*\left({\bm{x}^t_j}^*\right)\right)$ into $\mathcal{K}$.
		}
	
		Provide $\mathcal{K}$ to learners.
		
		\BlankLine
		\textbf{The learner} updates $f^t$ based on received $\mathcal{K}$:
		
		$f^t\gets f^t-\eta^t\mathcal{G}(\mathcal{L};f^t;\mathcal{K})$.
		\BlankLine
		Set $t\gets t+1$.
		}
\end{algorithm}

Allowing more examples to be fed, \ie, feeding a pack of teaching examples instead of a single one at each iteration, we present the Greedy-$k$ Functional Teaching (GFT-$k$) as a heuristic. Given a nonparametric target model $f^*$, GFT-$k$ is to pick $k$ examples satisfying 
	\begin{eqnarray}
		\resizebox{.9\hsize}{!}{$\left({\bm{x}^t_j}^*=\underset{\bm{x}^t_i\in\mathcal{X}-\{{\bm{x}^t_i}^*\}_{i=1}^{j-1}}{\arg\max}\left\|\left.\frac{\partial\mathcal{L}}{\partial f}\right|_{f^t(\bm{x}^t_i),y^t_i}K \left(\bm{x}^t_i,\cdot\right)\right\|_\mathcal{H},{y^t_j}^*=f^*\left({\bm{x}^t_j}^*\right)\right)$}
	\end{eqnarray}
	as the pack of optimal examples to learners at $t$-th iteration, $t=0,1,\dots,\text{ITD}_\text{GFT}$ and $j=1,\dots,k$.

The hyper parameter $k$ can take the form of either an integer counting the number of examples, where $k\in\mathbb{N}$, or a decimal representing the ratio of the pack to the whole pool, where $k\in[0,1]$. The pseudo code for RFT, GFT-1, and GFT-$k$ is given in Algorithm~\ref{aaft} which encapsulates these algorithms.

\begin{rem}
	\label{pbt}
	For the pool-based teacher who can only provide teaching examples from a pool $\mathcal{P}\subsetneq\mathcal{X}$, RFT and GFT could still work by replacing $\mathcal{X}$ by $\mathcal{P}$. However, $f^t$ might converge to the suboptimal ${f^*}'$ when the optimal examples ${\bm{x}^t}^*\in\mathcal{X}-\mathcal{P}$ and therefore the pool-based teacher cannot provide them to learners.
\end{rem}

\subsection{Analysis of Iterative Teaching Dimension} \label{ftca}

We begin with iterative teaching dimension analysis of RFT under the assumptions \cite{shen2020sinkhorn} on $\mathcal{L}$ and the kernel function $K(\bm{x},\bm{x}')\in\mathcal{H}$ as below.
\begin{assu}
	\label{lls}
	The loss function $\mathcal{L}(f)$ is $L_\mathcal{L}$-Lipschitz smooth, \ie, $\forall f,f'\in\mathcal{H}$ and $\bm{x}\in\mathcal{X}$ $$\left|E_{\bm{x}}\left[\nabla_f \mathcal{L}(f)\right]-E_{\bm{x}}\left[\nabla_f\mathcal{L}(f')\right]\right|\leq L_\mathcal{L} \left|E_{\bm{x}}\left[f\right]-E_{\bm{x}}\left[f'\right]\right|, $$ where $L_\mathcal{L}\geq0$ is a constant.
\end{assu}

\begin{assu}
	\label{bkf}
	The kernel function $K(\bm{x},\bm{x}')\in\mathcal{H}$ is bounded, \ie, $\forall \bm{x},\bm{x}' \in\mathcal{X},\,K(\bm{x},\bm{x}')\leq M_K$, where $M_K\geq0$ is a constant.
\end{assu}

Recall the definition of the evaluation functional and Fréchet derivative in Definition~\ref{efl} and \ref{defn1}, respectively, we further introduce a discrepancy \cite{shen2020sinkhorn} to quantify the inconsistency between $f^t$ and $f^*$ before theoretical analysis.
\begin{defn}
	\label{disc}
	The discrepancy of iterative teaching between $f^t$ and $f^*$ at $\bm{x}^t$ is defined as
	\begin{equation}
		\label{itd}
		\mathbb{S}_{\mathcal{L}}(f^t;{\bm{x}^t})\coloneqq\left|E_{\bm{x}^t}\nabla_f \mathcal{L}(f^t,f^*)\right|^2.
	\end{equation}
\end{defn}

For succinctness, we rewrite Eq.~\ref{itd} as $\mathbb{S}_{\mathcal{L}}(f^t;{\bm{x}^t})=\left|E_{\bm{x}^t}\nabla_f \mathcal{L}(f^t)\right|^2$ by omitting given $f^*$. One can observe that $\mathbb{S}_{\mathcal{L}}(f^t;{\bm{x}^t})$ decreases as $f^t$ approaches $f^*$, thus it can track the convergence state of functional teaching algorithms and measure the per-iteration improvement about $f^t$ towards $f^*$. Interestingly, the discrepancy of iterative teaching shares a close connection with the Fisher information \cite{rissanen1996fisher, schervish2012theory}. Note that $\left|E_{\bm{x}^t}\nabla_f \mathcal{L}(f^t)\right|^2$ can be equivalently written as $E_{\bm{x}}\left(\nabla_f \mathcal{L}(f)\right)^2$. Focus on arithmetic mean rather than point evaluation of $\left(\nabla_f \mathcal{L}(f)\right)^2$, then replacing evaluation functional operator by expectation operator, we have $\mathbb{E}_{\bm{x}\sim\mathbb{P}(\bm{x})}\{\left(\nabla_f \mathcal{L}(\bm{x};f)\right)^2\}$, which can be viewed as a nonparametric Fisher information for convex loss function. Let $f$ degenerate into the unknown parameter $\theta$ and $\mathcal{L}$ be the natural logarithm of the likelihood function $\ell(\bm{x};\theta)$, we have $\mathbb{E}_{\bm{x}\sim\mathbb{P}(\bm{x})}\{\left(\nabla_\theta \log\ell(\bm{x};\theta)\right)^2\}$. Therefore, nonparametric Fisher information for convex loss function can be viewed as a kind of generalized Fisher information, which extends the natural logarithm of likelihood function to a convex loss function and the unknown parameter to a general mapping. More discussion is in Appendix~\ref{nlfi}.

\textbf{Random Functional Teaching.} Recall the teaching settings (Eq.~\ref{la}, Eq.~\ref{opta}), we analyze per-iteration reduction w.r.t. $\mathcal{L}$.

\begin{lem}(Sufficient Descent for RFT)
	\label{sdpft}
	Under Assumption~\ref{lls} and \ref{bkf}, if $\eta^t\leq 1/(2L_\mathcal{L}\cdot M_K)$, RFT teachers can reduce the loss $\mathcal{L}$:
	\begin{equation}
		\mathcal{L}(f^{t+1})-\mathcal{L}(f^t)\leq-\eta^t/2\cdot\mathbb{S}_{\mathcal{L}}(f^t;{\bm{x}^t}).
	\end{equation}
\end{lem}

Proof of the Lemma~\ref{sdpft} is in Appendix~\ref{psdpft}. Before the convergence of RFT algorithm, the decrease of $\mathcal{L}$ has a negative upper bound expressed by $\mathbb{S}_{\mathcal{L}}(f^t;{\bm{x}^t})$, which is determined by learning rate $\eta^t$, loss $\mathcal{L}$, mastery degree $f^t$ and teaching example $\bm{x}^t$. One can see that these four factors are independent so they affect per-iteration reduction of $\mathcal{L}$ independently. Therefore, even though teachers fail to observe all factors under the gray-box setting, they can also assume that unknown factors are fixed, and optimize example feeding based on tracked $f^t$ to steepen gradients. This is consistent with the motivation of GFT deriving from Proposition~\ref{gs} and Theorem~\ref{optthm}. 

\begin{thm}(Convergence for RFT)
	\label{cpft}
	Suppose the model of learners is initialized with $f^0\in\mathcal{H}$ and returns $f^t\in\mathcal{H}$ after $t$ iterations, we have the upper bound of minimal $\mathbb{S}_{\mathcal{L}}(f^t;{\bm{x}^t})$:
	\begin{eqnarray}\label{eqcpft}
		\min_t\mathbb{S}_{\mathcal{L}}(f^t;{\bm{x}^t}) \leq2\mathcal{L}(f^0)/\left(\tilde{\eta}t\right),
	\end{eqnarray}
	where $0<\tilde{\eta}=\underset{t}{\min}\,\eta^t\leq \frac{1}{2L_\mathcal{L}\cdot M_K}$.
\end{thm}

The proof of the Theorem~\ref{cpft} is given in Appendix~\ref{pcpft}. It follows from Eq.~\ref{eqcpft} that the upper bound of minimal $\mathbb{S}_{\mathcal{L}}(f^t;{\bm{x}^t})$ converges at the rate of $\mathcal{O}(1/t)$ and $\min_t\mathbb{S}_{\mathcal{L}}(f^t;{\bm{x}^t})\to0$ as $t\to\infty$, which means it needs $\mathcal{O}(2\mathcal{L}(f^0)/\left(\tilde{\eta}\epsilon\right))$ iterations for RFT to achieve a stationary point with constant $\epsilon>0$. Therefore, we conclude that ITD of RFT is $\mathcal{O}(2\mathcal{L}(f^0)/\left(\tilde{\eta}\epsilon\right)) < \infty$, which suggests feasibility of our extension from the parametric IMT to nonparametric IMT.

\textbf{Greedy Functional Teaching.} Compared to RFT, GFT provably enjoys a faster convergence rate and needs fewer iterations to converge, \ie, lower ITD.

\begin{lem}(Sufficient Descent for GFT)
	\label{sdaft}
	Under Assumption~\ref{lls} and \ref{bkf}, if $\eta^t\leq 1/(2L_\mathcal{L}\cdot M_K)$, GFT teachers can reduce the loss $\mathcal{L}$ at a faster speed:
	\begin{eqnarray}
		\mathcal{L}(f^{t+1})-\mathcal{L}(f^t)&\leq&-\eta^t/2\cdot\mathbb{S}_{\mathcal{L}}(f^t;{\bm{x}^t}^*)\nonumber\\
		&\leq&-\eta^t/2\cdot\mathbb{S}_{\mathcal{L}}(f^t;{\bm{x}^t}).
	\end{eqnarray}
\end{lem}

The proof of the Lemma~\ref{sdaft} is presented in Appendix~\ref{psdaft}. One can observe that per-round improvement of GFT has a tighter bound than that of RFT. The reason is that with a greedy strategy GFT elaborately selects examples by maximizing norm of difference between current and target models, such that learners improve $f^t$ with a steeper step forward $f^*$ in per iteration. Such tighter bound approves the efficiency of GFT theoretically.

\begin{thm}(Convergence for GFT)
	\label{caft}
	Suppose the model of learners is initialized with $f^0\in\mathcal{H}$ and returns $f^t\in\mathcal{H}$ after $t$ iterations, we have the upper bound of minimal $\mathbb{S}_{\mathcal{L}}(f^t;{\bm{x}^j}^*)$:
	\begin{eqnarray}
		\min_j\mathbb{S}_{\mathcal{L}}(f^j;{\bm{x}^j}^*)\leq\frac{2}{\tilde{\eta}\psi(t)}\mathcal{L}(f^0),
	\end{eqnarray}
	where $0<\tilde{\eta}=\underset{t}{\min}\,\eta^t\leq \frac{1}{2L_\mathcal{L}\cdot M_K}$, $\psi(t)=\sum_{j=0}^{t-1}\gamma^j$ and $\gamma^j = 
	\frac{\mathbb{S}_{\mathcal{L}}(f^j;{\bm{x}^j})}{\mathbb{S}_{\mathcal{L}}(f^j;{\bm{x}^j}^*)}\in(0,1]$ named greedy ratio.
\end{thm}

\begin{figure*}[t]
	\subfigbottomskip=-4pt
	\subfigcapskip=-4pt
	\centering
	\subfigure[Regression: Teaching a Posterior Probability Density Function.]{\includegraphics[width=\linewidth]{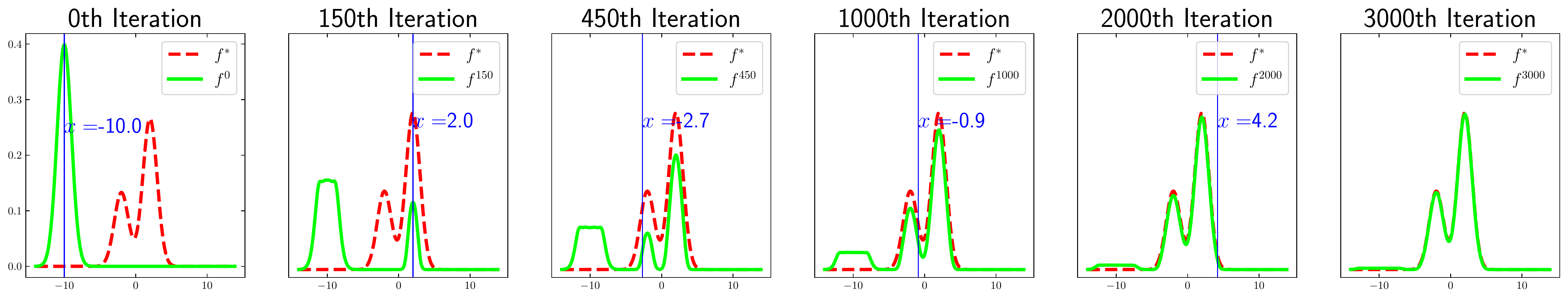}}
	\subfigure[Classification: Teaching a Nonlinear Decision Boundary. ]{\includegraphics[width=\linewidth]{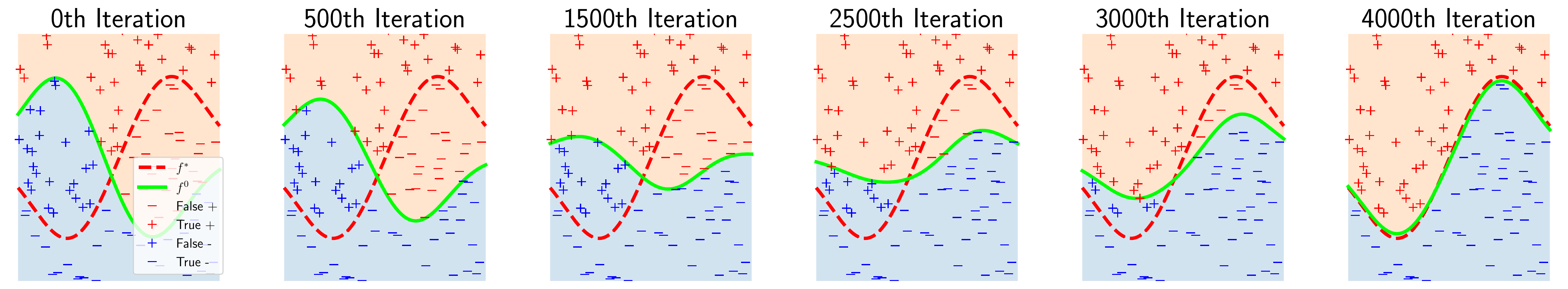}}
	\vskip 0.35pt
	\caption{GFT for nonparametric regression and classification teaching problems. (a): The red dashed lines are $f^*$ and the solid lime lines are $f^t$ at different iteration of GFT. Selected examples are pointed out by blue vertical lines. (b): The red dashed lines are $f^*$ when $f^0$ is represented by the edge between blue and orange regions. $x_1$ and $x_2$ are corresponded to $x$ and $y$ axis, respectively. (a)-(b) present the nonparametric teaching ability of helping the learner converge to $f^*$ even from a terrible initial $f^0$ (without overlap with $f^*$).}
	\label{synt}
	\vskip 0.05in
\end{figure*}

The proof of the Theorem~\ref{caft} is given in Appendix~\ref{pcaft}. Greedy ratio measures the per-iteration reduction difference between RFT and GFT, and $\psi(t)$ thereby denotes the cumulative difference, \ie superiority of GFT compared to RFT. Intuitively, GFT is strikingly efficient than RFT at beginning and greedy ration is close to $0$. As teaching goes on, such divergence vanishes gradually, then greedy ration increasingly close to $1$. For $\underset{{t\to \infty}}{\lim}\gamma^t\to1$, we must have $\underset{{t\to \infty}}{\lim}\psi(t)\to\infty$. Since $\psi(t)\leq t$, one can obtain
\begin{equation}
	\frac{2}{\tilde{\eta}\psi(t)}\mathcal{L}(f^0)\geq\frac{2}{\tilde{\eta}t}\mathcal{L}(f^0),
\end{equation}
which means $\min_j\mathbb{S}_{\mathcal{L}}(f^j;{\bm{x}^j}^*)$ has a higher upper bound than $\min_j\mathbb{S}_{\mathcal{L}}(f^j;{\bm{x}^j})$. In another word, GFT holds a lower ITD $\mathcal{O}(\psi(\frac{2\mathcal{L}(f^0)}{\tilde{\eta}\epsilon}))$ for convergence.

\begin{rem}
	Computation complexity. The major computational cost comes from gradient calculation, which could be sped up via parallelization provided in GFT-$k$. Besides, when the size of an example pool is $n$, Kernel Operation (KO) and example selection for GFT cost $\mathcal{O}(n^2)$ and $\mathcal{O}(n)$, respectively. In large-scale problem, cost of GFT could be saved by implementing it in sub-sampled support of $f^*$ \cite{politis1999subsampling} and cost of KO could be cut down by a random feature expansion of the kernel \cite{rahimi2007random, liu2017stein}.
\end{rem}

\section{Experiments and Results}
\vspace{.5mm}

We test our RFT and GFT on both synthetic and real-world data, on which we find these two algorithms present satisfactory capability to tackle nonparametric teaching tasks. Without particular emphasis, experiments are implemented under the synthesis-based teacher setting where the teacher can provide any examples to learners and the knowledge domain is complete. Some detailed settings and extended experiments are given in Appendix~\ref{de}, \ref{ee}.

\vspace{2.5mm}

\textbf{Synthetic 1D Gaussian Mixture.} Consider a nonparametric Bayesian inference problem. The target model is specified as the posterior probability density function (PDF) set to be $f^* = 1/3\mathcal{N}(x;-2,1)+2/3\mathcal{N}(x;2,1)$, where we denote the PDF of a normal distribution with mean $\mu$ and standard deviation $\sigma$ as $\mathcal{N}(x;\mu,\sigma)$. We assume $f^0$ for the learner is initialized as $f^0=\mathcal{N}(x;-10,1)$. This is a challenging regression teaching problem since $f^0$ and $f^*$ is far apart (almost without overlap). (a) in Fig.~\ref{synt} shows that $f^0$ is guided by GFT to evolve towards $f^*$ directly. It can be found that in spite of obvious difference between $f^*$ and $f^0$, our GFT can smooth the mode of $f^0$ where is flatten in $f^*$ and sharpen $f^0$ towards the mode of $f^*$ via searching $x$ with maximal $\left|f^*(x)-f^t(x)\right|$ and feeding it to the learner.

\begin{figure}[t]
	\subfigbottomskip=-6pt
	\subfigcapskip=-4pt
	\centering
	\subfigure[GFT-1]{\includegraphics[width=\linewidth]{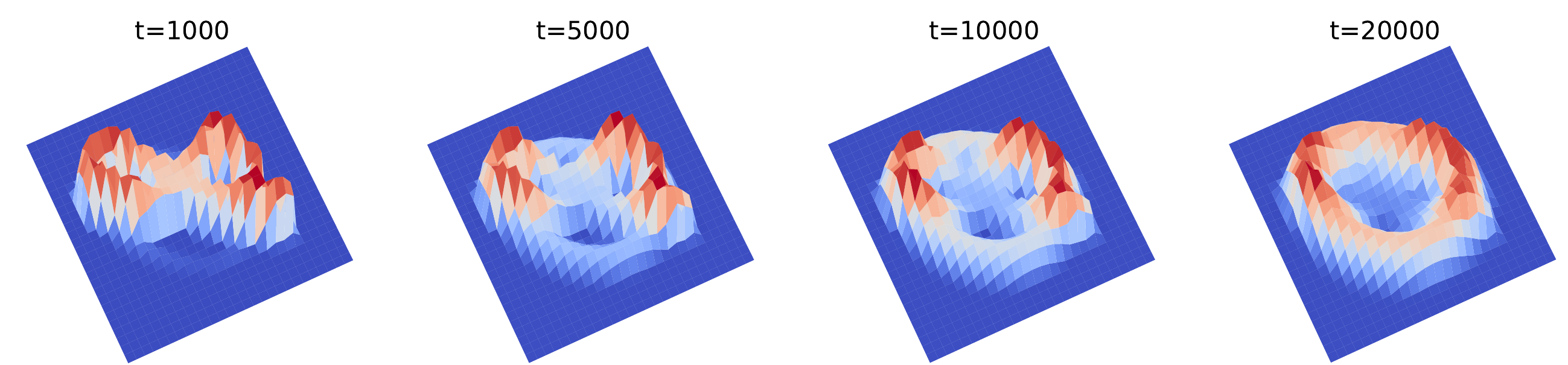}}
	\subfigure[Pool GFT-1]{\includegraphics[width=\linewidth]{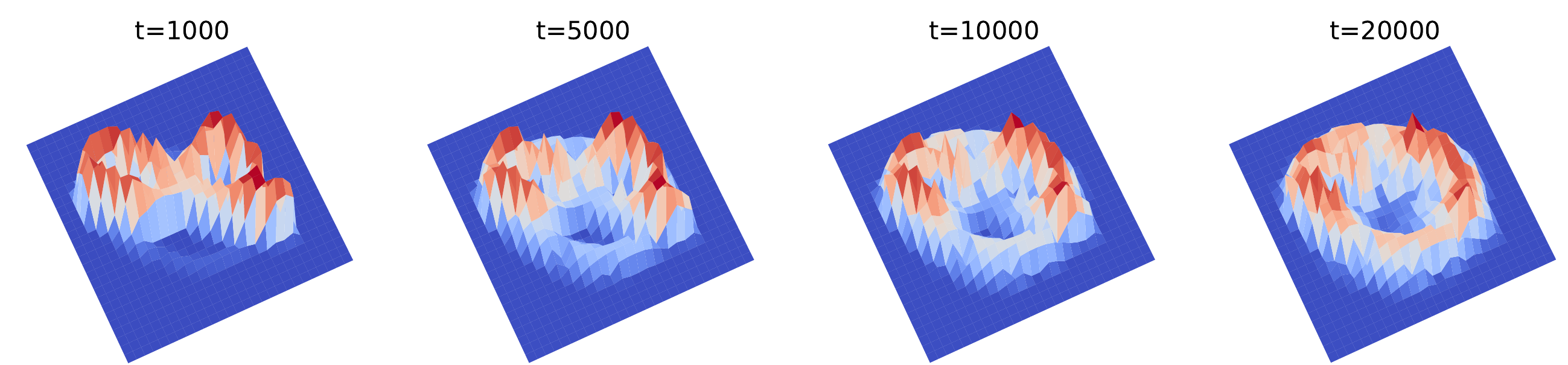}}
	\subfigure[GFT-1 with Alternative]{\includegraphics[width=\linewidth]{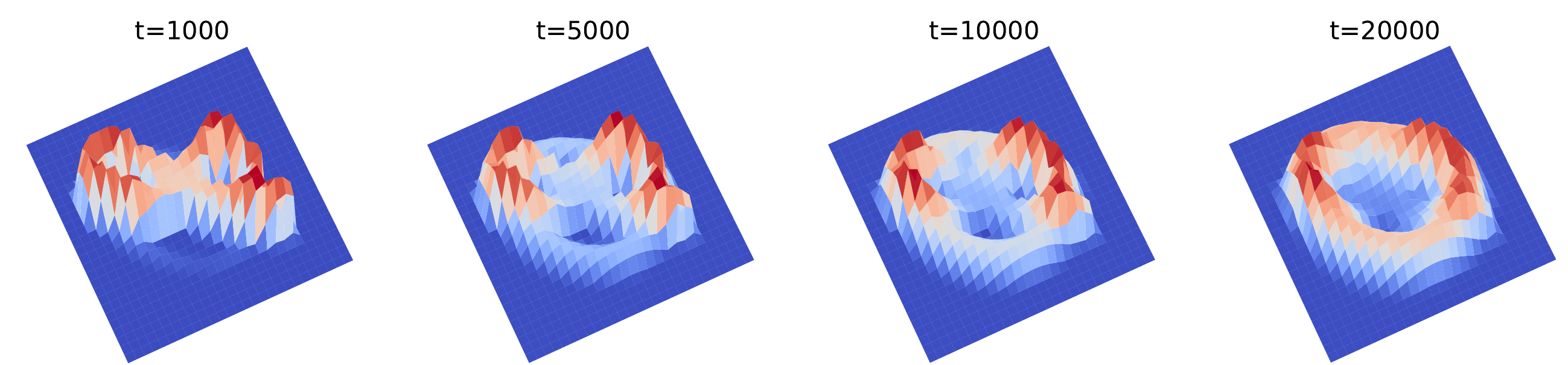}}
	\vskip 0.1pt
	\caption{Nonparametric teaching for correcting 8 towards 0. (a): evolution of $f^t$ with GFT-1 algorithm. (b): $f^t$ for GFT-1 under the pool-based teacher. (c): $f^t$ for GFT-1 when occasionally teaching with O. GFT-1 presents satisfied nonparametric teaching capability in these different scenarios.}
	\vskip -0.1in
	\label{np80}
\end{figure}

\textbf{Synthetic 2D Classification.} For a 2D nonparametric classification problem, out of convenience for visualization, the target model is set to be $f^*(x_1,x_2)=x_2-\exp\left(\frac{x_1-0.5}{0.5}\right)^2+\exp\left(\frac{x_1+0.5}{0.5}\right)^2$, where $x_i$ represents feature $i$, $i=1,2$. Then, let $f^*(x_1,x_2)=0$, we can rewrite it as $x_2=\exp\left(\frac{x_1-0.5}{0.5}\right)^2-\exp\left(\frac{x_1+0.5}{0.5}\right)^2$ and visualize the decision boundary in a 2D figure. $f^0$ is set to be $f^0=x_2+\exp\left(\frac{x_1-0.3}{0.5}\right)^2-\exp\left(\frac{x_1+0.6}{0.5}\right)^2$, from which we have $x_2=\exp\left(\frac{x_1+0.6}{0.5}\right)^2-\exp\left(\frac{x_1-0.3}{0.5}\right)^2$. (b) in Fig.~\ref{synt} presents how GFT corrects the inappropriate decision boundary $f^0$ towards $f^*$. It can be observed that for a more general function more than PDF, our GFT is also able to amend a bad initialization $f^0$ towards $f^*$.

More experiments applying RFT and GFT to teach parameterized target models are given in Appendix~\ref{ee}, which shows parametric adaptation of RFT and GFT.

\textbf{The digit Correction.} Consider a digit (MNIST \cite{lecun1998mnist}) teaching instance, one can image a digit figure as a surface in 3D space where $z$ axis is the gray level and $x,y$ axes represent the pixel location. Obviously such complexity surface cannot be identified by a parameter, thus is beyond the capabilities of parametric algorithms \cite{liu2017iterative}. Initially, the teacher would ask an infant (the learner)\textit{ what is digit 0 }($f^*$)? He would provide a self-convinced but wrong answer as digit 8 ($f^0$) to the teacher. Based on such a feedback, the teacher would correct $f^0$ towards $f^*$ via feeding examples (fundamentally is gray value with pixel location). After many rounds of teaching and learning, the learner would evolve its $f$ from incorrect $f^0$ to ambiguous $f^t$ and final correct $f^*$, which shares similarity with the process when human beings learn new items \cite{bengio2009curriculum}. We visualize above procedure of our GFT-1 teacher in Fig.~\ref{np80} (a).

\vspace{-1mm}
\begin{figure}[t]
	\subfigbottomskip=-4pt
	\subfigcapskip=-3pt
	\centering
	\subfigure[The Digit Correction]{\includegraphics[width=0.49\linewidth]{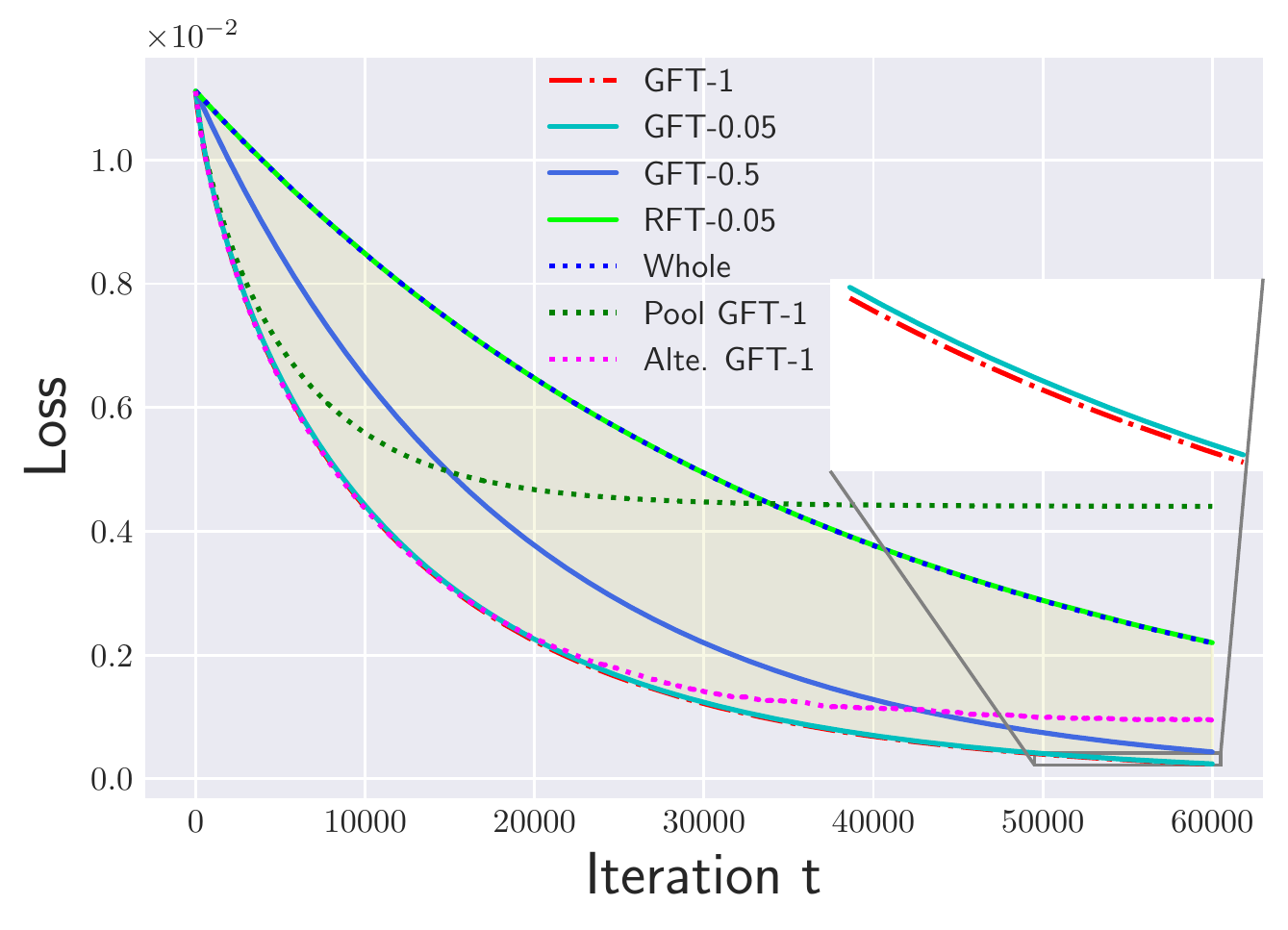}}
	\subfigure[The Cheetah Impartation]{\includegraphics[width=0.49\linewidth]{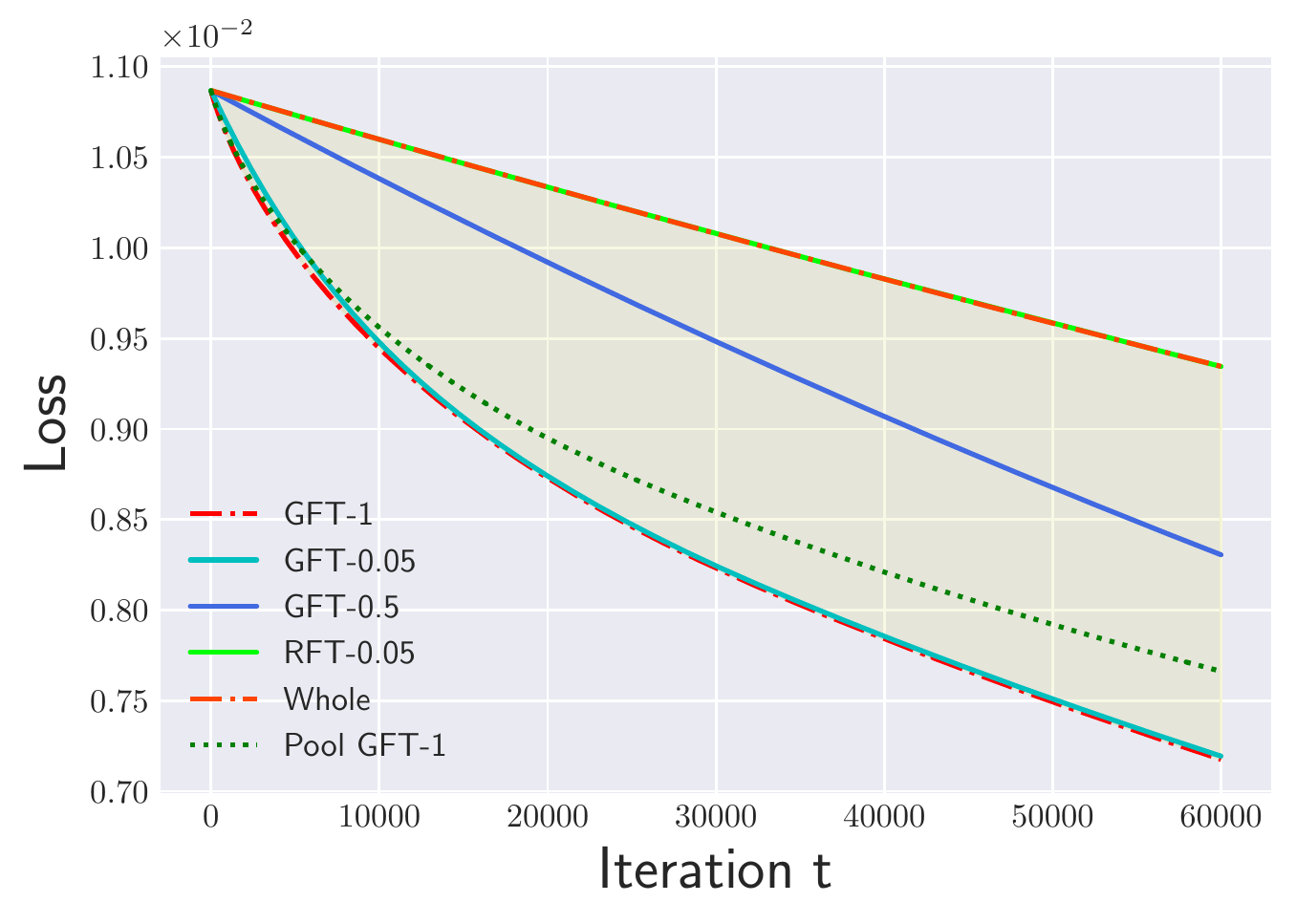}}
	\vskip 0.1pt
	\caption{Comparison of convergence performance for RFT and GFT. The legend of GFT-1 for pool-based teaching is Pool GFT-1 when that for alternative teaching is Alte. GFT-1. The legend, Whole means the teacher provides all pixels to the learner.}
	\vskip -0.1in
	\label{tl}
\end{figure}

Consider practical pool-based teacher scenario (introduced in Remark~\ref{pbt}). We randomly set that $80\%$ pixels are available to the pool-based teacher as $\mathcal{P}$. Fig.~\ref{np80} (b) shows that our GFT-1 is also effective while $f^t$ cannot converge to $f^*$ due to the limited knowledge domain of the pool-based teacher. 

A more interesting case is alternative teaching. Specifically, digit 0 is well-known for the teacher, but lack of Kids Picture Dictionary of 0 at hand he cannot provide wanted teaching examples. Alternatively, notice on similar topological structure between digit 0 and character O (EMNIST from \cite{cohen2017emnist}), it is natural to take O as teaching examples. We set the probability of teaching with O as $0.2$ in each iteration to test GFT-1. As expected, Fig.~\ref{np80} (c) shows that GFT-1 also adapts to the alternative teaching with satisfied performance. It demonstrates generalizability of GFT-1 as it can be applied in more practical scenarios where only alternative with similar topological structure is accessible. This interesting property may present an intimate connection between our work and transfer learning \cite{pan2009survey}, domain adaptation \cite{daume2006domain}.

Fig.~\ref{tl} (a) presents the convergence performance for RFT and GFT under different settings. The yellow region is marked for GFT-$k$, $k\in(0,1)$. We see that the loss of GFT declines more dramatically than that of RFT, and it converges to sub-optimal $f$ under the pool-based teacher or alternative teaching scenarios. We leave comparison between RFT and GFT of concrete images like Fig.~\ref{np80} in Appendix~\ref{de} Fig.~\ref{np80m}.

\begin{figure}[t]
	\subfigbottomskip=-4pt
	\subfigcapskip=-3pt
	\centering
	\subfigure[GFT-1]{\includegraphics[width=\linewidth]{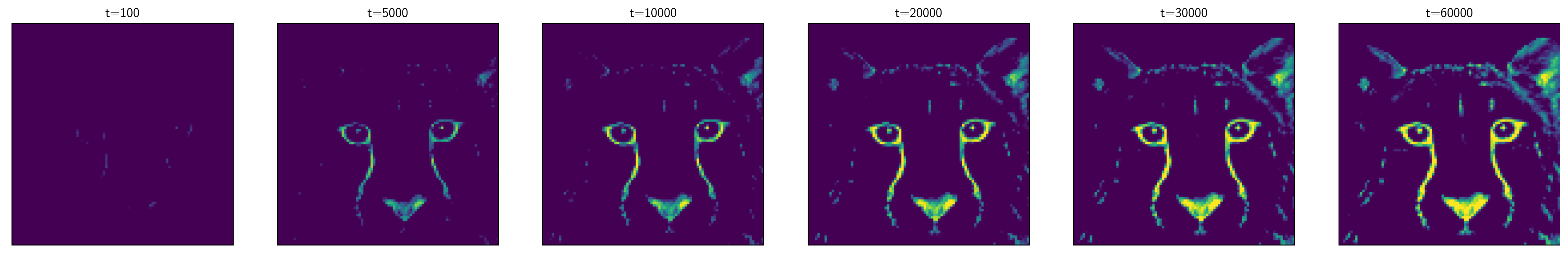}}
	\subfigure[Pool GFT-1]{\includegraphics[width=\linewidth]{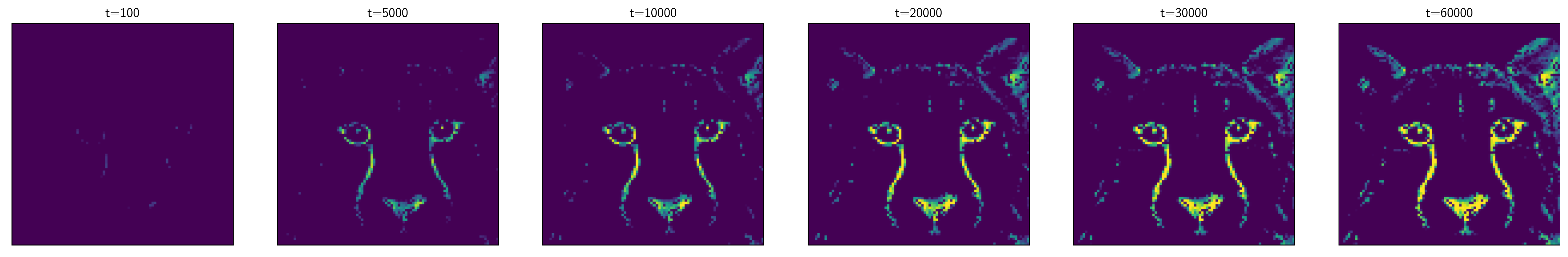}}
	\subfigure[GFT-0.05]{\includegraphics[width=\linewidth]{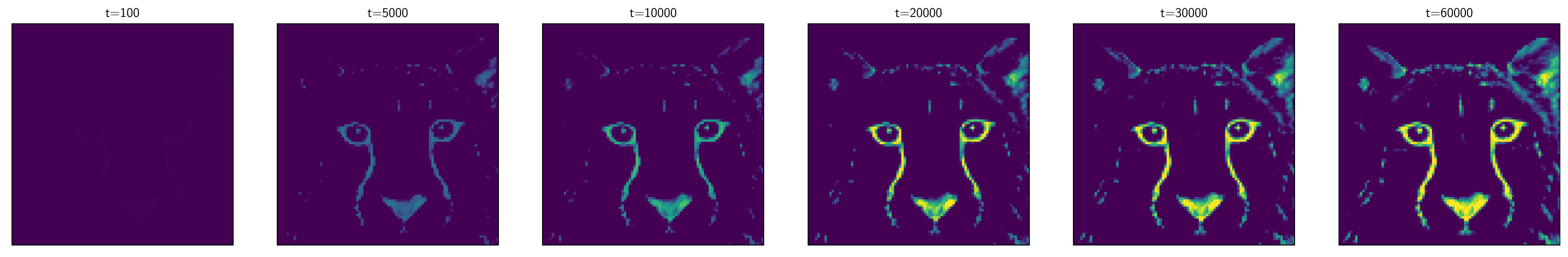}}
	\subfigure[GFT-0.5]{\includegraphics[width=\linewidth]{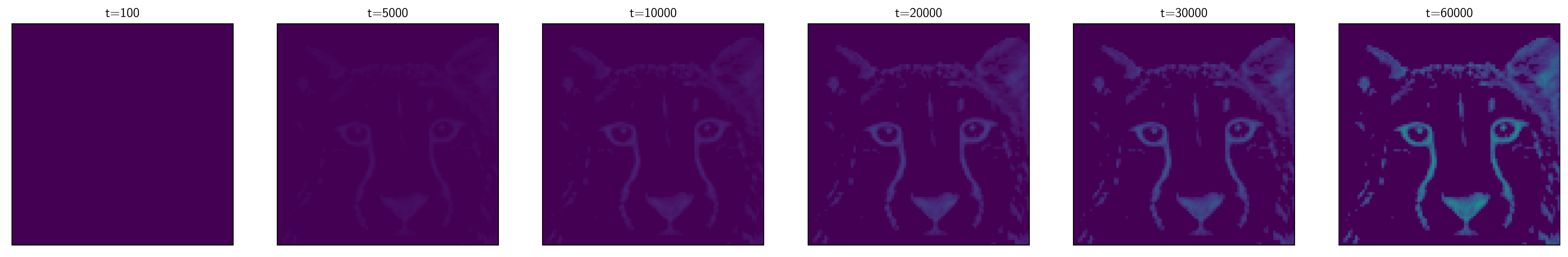}}
	\subfigure[RFT-0.05]{\includegraphics[width=\linewidth]{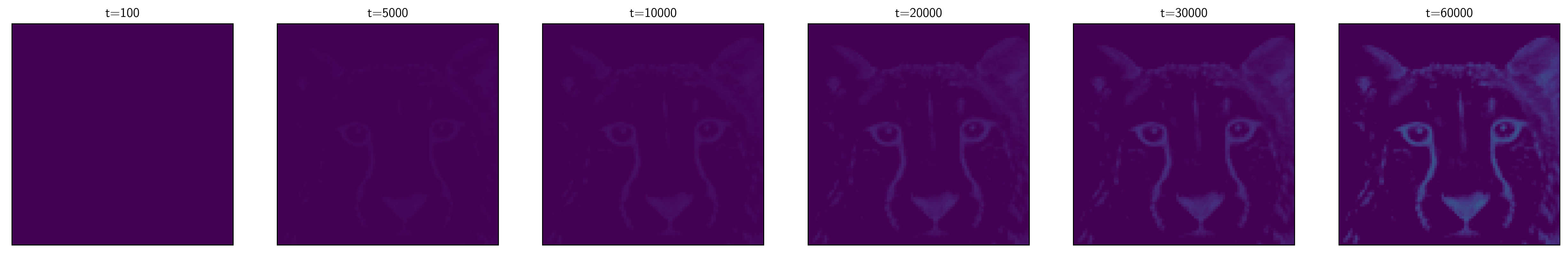}}
	\subfigure[Whole]{\includegraphics[width=\linewidth]{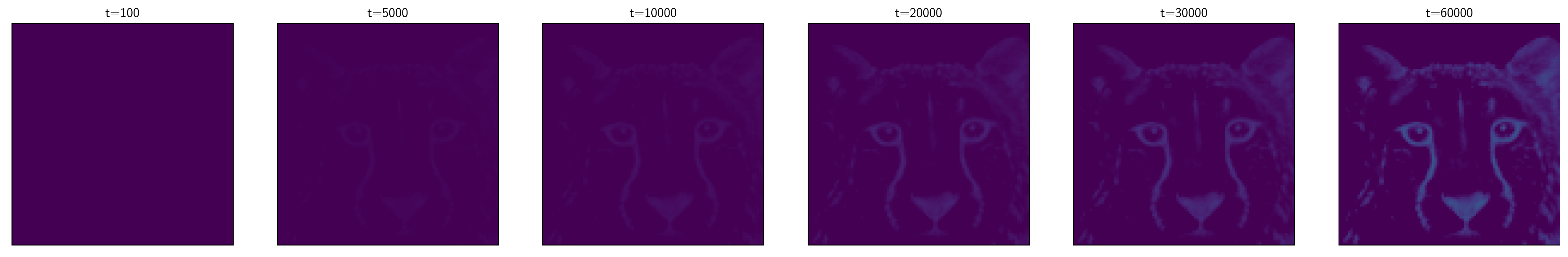}}
	\vskip -1.2pt
	\caption{Nonparametric teaching for imparting the cheetah. $f$ of GFT become clear significantly faster than RFT. Part of $f$ are not updated for pool-based teaching, so several dark discontinuity points can be found.}
	\label{nche}
	\vskip -0.1in
\end{figure}

\textbf{The cheetah impartation.} Different from correction tasks where the learner has a preliminary idea of $f^*$, the impartation problem focus on the learner with no idea about $f^*$. Concretely, when the teacher asks \textit{what is a cheetah} \cite{shen2020sinkhorn}, it would be a blank in the learner's mind. As a response, teacher would educate the learner about the cheetah in pixels viewpoint as breaking the whole concept down into smaller points brings better understanding. Fig.~\ref{nche} compares RFT and GFT under different settings by visualizing $f^t$ therein. We find that GFT is vastly better than RFT that has roughly the same performance as teaching with whole set \cite{bottou2010large}. Besides, GFT-1 tends to outperform other GFT algorithms, but fails to teach entirely $f^*$ under the pool-based setting.

We conclude from Fig.~\ref{tl} (b) that compared with GFT, RFT saves the cost of searching the optimal examples at the expense of slow convergence, and pool-based teaching also suffer from sub-optimization.

\section{Concluding Remarks}
In this paper, we study a general task, Nonparametric Iterative Machine Teaching (NIMT), which generalizes model space from a finite dimensional one to an infinite dimensional one. We are mainly concerned with the reproduce kernel Hilbert space in this paper. To tackle NIMT, we present a natural baseline algorithm named random functional teaching and propose a greedy one named greedy functional teaching. We theoretically prove that iterative teaching dimension of random functional teaching is $\mathcal{O}(2\mathcal{L}(f^0)/\left(\tilde{\eta}\epsilon\right))$ when greedy functional teaching has a lower iterative teaching dimension $\mathcal{O}(\psi(\frac{2\mathcal{L}(f^0)}{\tilde{\eta}\epsilon}))$ for convergence under mild assumptions. We experimentally demonstrate the efficiency of these two algorithms. Future directions could be more theoretical understanding on NIMT and more efficient functional teaching algorithms with better strategies for potential practical application in deep learning models.

\section*{Acknowledgements}
This work was supported in part by National Natural Science Foundation of China (Grant Number: 62206108), in part by Maritime AI Research Programme (SMI-2022-MTP-06) and AI Singapore OTTC Grant (AISG2-TC-2022-006), and in part by the Research Grants Council of the Hong Kong Special Administrative Region (Grant 16200021).

\bibliography{main.bib}

\begin{thebibliography}{83}
\providecommand{\natexlab}[1]{#1}
\providecommand{\url}[1]{\texttt{#1}}
\expandafter\ifx\csname urlstyle\endcsname\relax
  \providecommand{\doi}[1]{doi: #1}\else
  \providecommand{\doi}{doi: \begingroup \urlstyle{rm}\Url}\fi

\bibitem[Adams \& Fournier(2003)Adams and Fournier]{adams2003sobolev}
Adams, R.~A. and Fournier, J.~J.
\newblock \emph{Sobolev spaces}.
\newblock Elsevier, 2003.

\bibitem[Alfeld et~al.(2016)Alfeld, Zhu, and Barford]{alfeld2016data}
Alfeld, S., Zhu, X., and Barford, P.
\newblock Data poisoning attacks against autoregressive models.
\newblock In \emph{AAAI}, 2016.

\bibitem[Alfeld et~al.(2017)Alfeld, Zhu, and Barford]{alfeld2017explicit}
Alfeld, S., Zhu, X., and Barford, P.
\newblock Explicit defense actions against test-set attacks.
\newblock In \emph{AAAI}, 2017.

\bibitem[Arbel et~al.(2019)Arbel, Korba, Salim, and Gretton]{arbel2019maximum}
Arbel, M., Korba, A., Salim, A., and Gretton, A.
\newblock Maximum mean discrepancy gradient flow.
\newblock In \emph{NeurIPS}, 2019.

\bibitem[Becke(1988)]{becke1988density}
Becke, A.~D.
\newblock Density-functional exchange-energy approximation with correct
  asymptotic behavior.
\newblock \emph{Physical review A}, 38\penalty0 (6):\penalty0 3098, 1988.

\bibitem[Bengio et~al.(2009)Bengio, Louradour, Collobert, and
  Weston]{bengio2009curriculum}
Bengio, Y., Louradour, J., Collobert, R., and Weston, J.
\newblock Curriculum learning.
\newblock In \emph{ICML}, 2009.

\bibitem[Blei et~al.(2017)Blei, Kucukelbir, and McAuliffe]{blei2017variational}
Blei, D.~M., Kucukelbir, A., and McAuliffe, J.~D.
\newblock Variational inference: A review for statisticians.
\newblock \emph{Journal of the American statistical Association}, 112\penalty0
  (518):\penalty0 859--877, 2017.

\bibitem[Bottou(2010)]{bottou2010large}
Bottou, L.
\newblock Large-scale machine learning with stochastic gradient descent.
\newblock In \emph{Proceedings of COMPSTAT'2010}, pp.\  177--186. Springer,
  2010.

\bibitem[Boyd et~al.(2004)Boyd, Boyd, and Vandenberghe]{boyd2004convex}
Boyd, S., Boyd, S.~P., and Vandenberghe, L.
\newblock \emph{Convex optimization}.
\newblock Cambridge university press, 2004.

\bibitem[Chen et~al.(2018)Chen, Singla, Mac~Aodha, Perona, and
  Yue]{chen2018understanding}
Chen, Y., Singla, A., Mac~Aodha, O., Perona, P., and Yue, Y.
\newblock Understanding the role of adaptivity in machine teaching: The case of
  version space learners.
\newblock In \emph{NeurIPS}, 2018.

\bibitem[Cohen et~al.(2017)Cohen, Afshar, Tapson, and
  Van~Schaik]{cohen2017emnist}
Cohen, G., Afshar, S., Tapson, J., and Van~Schaik, A.
\newblock Emnist: Extending mnist to handwritten letters.
\newblock In \emph{IJCNN}, 2017.

\bibitem[Coleman(2012)]{coleman2012calculus}
Coleman, R.
\newblock \emph{Calculus on normed vector spaces}.
\newblock Springer Science \& Business Media, 2012.

\bibitem[Collins et~al.(2023)Collins, Bhatt, Liu, Piratla, Sucholutsky, Love,
  and Weller]{Collins2023HILLM}
Collins, K.~M., Bhatt, U., Liu, W., Piratla, V., Sucholutsky, I., Love, B., and
  Weller, A.
\newblock Human-in-the-loop mixup.
\newblock In \emph{UAI}, 2023.

\bibitem[Corder \& Foreman(2014)Corder and Foreman]{corder2014nonparametric}
Corder, G.~W. and Foreman, D.~I.
\newblock \emph{Nonparametric statistics: A step-by-step approach}.
\newblock John Wiley \& Sons, 2014.

\bibitem[Cormen et~al.(2022)Cormen, Leiserson, Rivest, and
  Stein]{cormen2022introduction}
Cormen, T.~H., Leiserson, C.~E., Rivest, R.~L., and Stein, C.
\newblock \emph{Introduction to algorithms}.
\newblock MIT press, 2022.

\bibitem[Daume~III \& Marcu(2006)Daume~III and Marcu]{daume2006domain}
Daume~III, H. and Marcu, D.
\newblock Domain adaptation for statistical classifiers.
\newblock \emph{Journal of artificial Intelligence research}, 26:\penalty0
  101--126, 2006.

\bibitem[Dvurechenskii et~al.(2018)Dvurechenskii, Dvinskikh, Gasnikov, Uribe,
  and Nedich]{dvurechenskii2018decentralize}
Dvurechenskii, P., Dvinskikh, D., Gasnikov, A., Uribe, C., and Nedich, A.
\newblock Decentralize and randomize: Faster algorithm for wasserstein
  barycenters.
\newblock In \emph{NeurIPS}, 2018.

\bibitem[Friedman(2001)]{friedman2001greedy}
Friedman, J.~H.
\newblock Greedy function approximation: a gradient boosting machine.
\newblock \emph{Annals of statistics}, pp.\  1189--1232, 2001.

\bibitem[Gelfand et~al.(2000)Gelfand, Silverman, et~al.]{gelfand2000calculus}
Gelfand, I.~M., Silverman, R.~A., et~al.
\newblock \emph{Calculus of variations}.
\newblock Courier Corporation, 2000.

\bibitem[Genevay et~al.(2016)Genevay, Cuturi, Peyr{\'e}, and
  Bach]{genevay2016stochastic}
Genevay, A., Cuturi, M., Peyr{\'e}, G., and Bach, F.
\newblock Stochastic optimization for large-scale optimal transport.
\newblock In \emph{NIPS}, 2016.

\bibitem[Goldman \& Kearns(1995)Goldman and Kearns]{goldman1995complexity}
Goldman, S.~A. and Kearns, M.~J.
\newblock On the complexity of teaching.
\newblock \emph{Journal of Computer and System Sciences}, 50\penalty0
  (1):\penalty0 20--31, 1995.

\bibitem[Hardt et~al.(2016)Hardt, Recht, and Singer]{hardt2016train}
Hardt, M., Recht, B., and Singer, Y.
\newblock Train faster, generalize better: Stability of stochastic gradient
  descent.
\newblock In \emph{ICML}, 2016.

\bibitem[He et~al.(2016)He, Zhang, Ren, and Sun]{he2016deep}
He, K., Zhang, X., Ren, S., and Sun, J.
\newblock Deep residual learning for image recognition.
\newblock In \emph{CVPR}, 2016.

\bibitem[Hofmann et~al.(2008)Hofmann, Sch{\"o}lkopf, and
  Smola]{hofmann2008kernel}
Hofmann, T., Sch{\"o}lkopf, B., and Smola, A.~J.
\newblock Kernel methods in machine learning.
\newblock \emph{The annals of statistics}, 36\penalty0 (3):\penalty0
  1171--1220, 2008.

\bibitem[Hollander et~al.(2013)Hollander, Wolfe, and
  Chicken]{hollander2013nonparametric}
Hollander, M., Wolfe, D.~A., and Chicken, E.
\newblock \emph{Nonparametric statistical methods}.
\newblock John Wiley \& Sons, 2013.

\bibitem[Hunziker et~al.(2018)Hunziker, Chen, Mac~Aodha, Rodriguez, Krause,
  Perona, Yue, and Singla]{hunziker2018teaching}
Hunziker, A., Chen, Y., Mac~Aodha, O., Rodriguez, M.~G., Krause, A., Perona,
  P., Yue, Y., and Singla, A.
\newblock Teaching multiple concepts to a forgetful learner.
\newblock \emph{arXiv preprint arXiv:1805.08322}, 2018.

\bibitem[Kallenberg et~al.(1985)Kallenberg, Oosterhoff, and
  Schriever]{kallenberg1985number}
Kallenberg, W. C.~M., Oosterhoff, J., and Schriever, B.
\newblock The number of classes in chi-squared goodness-of-fit tests.
\newblock \emph{Journal of the American Statistical Association}, 80\penalty0
  (392):\penalty0 959--968, 1985.

\bibitem[Kamalaruban et~al.(2019)Kamalaruban, Devidze, Cevher, and
  Singla]{kamalaruban2019interactive}
Kamalaruban, P., Devidze, R., Cevher, V., and Singla, A.
\newblock Interactive teaching algorithms for inverse reinforcement learning.
\newblock \emph{arXiv preprint arXiv:1905.11867}, 2019.

\bibitem[Kumar et~al.(2021)Kumar, Zhang, Singla, and Chen]{kumar2021teaching}
Kumar, A., Zhang, H., Singla, A., and Chen, Y.
\newblock The teaching dimension of kernel perceptron.
\newblock In \emph{AISTATS}, 2021.

\bibitem[Lax(2002)]{lax2002functional}
Lax, P.~D.
\newblock \emph{Functional analysis}, volume~55.
\newblock John Wiley \& Sons, 2002.

\bibitem[LeCun(1998)]{lecun1998mnist}
LeCun, Y.
\newblock The mnist database of handwritten digits.
\newblock \emph{http://yann. lecun. com/exdb/mnist/}, 1998.

\bibitem[Lehmann \& Casella(2006)Lehmann and Casella]{lehmann2006theory}
Lehmann, E.~L. and Casella, G.
\newblock \emph{Theory of point estimation}.
\newblock Springer Science \& Business Media, 2006.

\bibitem[Lessard et~al.(2019)Lessard, Zhang, and Zhu]{lessard2019optimal}
Lessard, L., Zhang, X., and Zhu, X.
\newblock An optimal control approach to sequential machine teaching.
\newblock In \emph{AISTATS}, 2019.

\bibitem[Liu et~al.(2016)Liu, Zhu, and Ohannessian]{liu2016teaching}
Liu, J., Zhu, X., and Ohannessian, H.
\newblock The teaching dimension of linear learners.
\newblock In \emph{ICML}, 2016.

\bibitem[Liu(2017)]{liu2017stein}
Liu, Q.
\newblock Stein variational gradient descent as gradient flow.
\newblock In \emph{NIPS}, 2017.

\bibitem[Liu \& Wang(2016)Liu and Wang]{liu2016stein}
Liu, Q. and Wang, D.
\newblock Stein variational gradient descent: A general purpose bayesian
  inference algorithm.
\newblock In \emph{NIPS}, 2016.

\bibitem[Liu et~al.(2017)Liu, Dai, Humayun, Tay, Yu, Smith, Rehg, and
  Song]{liu2017iterative}
Liu, W., Dai, B., Humayun, A., Tay, C., Yu, C., Smith, L.~B., Rehg, J.~M., and
  Song, L.
\newblock Iterative machine teaching.
\newblock In \emph{ICML}, 2017.

\bibitem[Liu et~al.(2018)Liu, Dai, Li, Liu, Rehg, and Song]{liu2018towards}
Liu, W., Dai, B., Li, X., Liu, Z., Rehg, J., and Song, L.
\newblock Towards black-box iterative machine teaching.
\newblock In \emph{ICML}, 2018.

\bibitem[Liu et~al.(2021)Liu, Liu, Wang, Paull, Schölkopf, and
  Weller]{Liu2021LAST}
Liu, W., Liu, Z., Wang, H., Paull, L., Schölkopf, B., and Weller, A.
\newblock Iterative teaching by label synthesis.
\newblock In \emph{NeurIPS}, 2021.

\bibitem[Lutwak et~al.(2012)Lutwak, Lv, Yang, and Zhang]{lutwak2012extensions}
Lutwak, E., Lv, S., Yang, D., and Zhang, G.
\newblock Extensions of fisher information and stam's inequality.
\newblock \emph{IEEE transactions on information theory}, 58\penalty0
  (3):\penalty0 1319--1327, 2012.

\bibitem[Lv(2017)]{lv2017general}
Lv, S.
\newblock General fisher information matrices of a random vector.
\newblock \emph{Advances in Applied Mathematics}, 89:\penalty0 18--40, 2017.

\bibitem[Ma et~al.(2019)Ma, Zhang, Sun, and Zhu]{ma2019policy}
Ma, Y., Zhang, X., Sun, W., and Zhu, J.
\newblock Policy poisoning in batch reinforcement learning and control.
\newblock In \emph{NeurIPS}, 2019.

\bibitem[Mansouri et~al.(2019)Mansouri, Chen, Vartanian, Zhu, and
  Singla]{mansouri2019preference}
Mansouri, F., Chen, Y., Vartanian, A., Zhu, J., and Singla, A.
\newblock Preference-based batch and sequential teaching: Towards a unified
  view of models.
\newblock In \emph{NeurIPS}, 2019.

\bibitem[Mason et~al.(1999{\natexlab{a}})Mason, Baxter, Bartlett, and
  Frean]{mason1999boosting}
Mason, L., Baxter, J., Bartlett, P., and Frean, M.
\newblock Boosting algorithms as gradient descent.
\newblock In \emph{NIPS}, 1999{\natexlab{a}}.

\bibitem[Mason et~al.(1999{\natexlab{b}})Mason, Baxter, Bartlett, Frean,
  et~al.]{mason1999functional}
Mason, L., Baxter, J., Bartlett, P.~L., Frean, M., et~al.
\newblock Functional gradient techniques for combining hypotheses.
\newblock In \emph{NIPS}, 1999{\natexlab{b}}.

\bibitem[Mroueh et~al.(2019)Mroueh, Sercu, and Raj]{mroueh2019sobolev}
Mroueh, Y., Sercu, T., and Raj, A.
\newblock Sobolev descent.
\newblock In \emph{AISTATS}, 2019.

\bibitem[Narici \& Beckenstein(2010)Narici and
  Beckenstein]{narici2010topological}
Narici, L. and Beckenstein, E.
\newblock \emph{Topological vector spaces}.
\newblock Chapman and Hall/CRC, 2010.

\bibitem[Nitanda \& Suzuki(2018)Nitanda and Suzuki]{nitanda2018functional}
Nitanda, A. and Suzuki, T.
\newblock Functional gradient boosting based on residual network perception.
\newblock In \emph{ICML}, 2018.

\bibitem[Nitanda \& Suzuki(2020)Nitanda and Suzuki]{nitanda2020functional}
Nitanda, A. and Suzuki, T.
\newblock Functional gradient boosting for learning residual-like networks with
  statistical guarantees.
\newblock In \emph{AISTATS}, 2020.

\bibitem[Noguchi(1994)]{noguchi1994invariant}
Noguchi, M.
\newblock Invariant fisher information.
\newblock \emph{Differential Geometry and its Applications}, 4\penalty0
  (2):\penalty0 179--199, 1994.

\bibitem[Pan \& Yang(2009)Pan and Yang]{pan2009survey}
Pan, S.~J. and Yang, Q.
\newblock A survey on transfer learning.
\newblock \emph{IEEE Transactions on knowledge and data engineering},
  22\penalty0 (10):\penalty0 1345--1359, 2009.

\bibitem[Peltola et~al.(2019)Peltola, {\c{C}}elikok, Daee, and
  Kaski]{peltola2019machine}
Peltola, T., {\c{C}}elikok, M.~M., Daee, P., and Kaski, S.
\newblock Machine teaching of active sequential learners.
\newblock In \emph{NeurIPS}, 2019.

\bibitem[Politis et~al.(1999)Politis, Romano, and Wolf]{politis1999subsampling}
Politis, D.~N., Romano, J.~P., and Wolf, M.
\newblock \emph{Subsampling}.
\newblock Springer Science \& Business Media, 1999.

\bibitem[Qian et~al.(2022)Qian, Liu, Su, Zhou, and Yu]{qian2022teaching}
Qian, H., Liu, X.-H., Su, C.-X., Zhou, A., and Yu, Y.
\newblock The teaching dimension of regularized kernel learners.
\newblock In \emph{ICML}, 2022.

\bibitem[Qiu et~al.(2022)Qiu, Liu, Xiao, Liu, Bhatt, Luo, Weller, and
  Sch{\"o}lkopf]{qiu2022iterative}
Qiu, Z., Liu, W., Xiao, T.~Z., Liu, Z., Bhatt, U., Luo, Y., Weller, A., and
  Sch{\"o}lkopf, B.
\newblock Iterative teaching by data hallucination.
\newblock In \emph{AISTATS}, 2022.

\bibitem[Rahimi \& Recht(2007)Rahimi and Recht]{rahimi2007random}
Rahimi, A. and Recht, B.
\newblock Random features for large-scale kernel machines.
\newblock \emph{Advances in neural information processing systems}, 20, 2007.

\bibitem[Rakhsha et~al.(2020)Rakhsha, Radanovic, Devidze, Zhu, and
  Singla]{rakhsha2020policy}
Rakhsha, A., Radanovic, G., Devidze, R., Zhu, X., and Singla, A.
\newblock Policy teaching via environment poisoning: Training-time adversarial
  attacks against reinforcement learning.
\newblock In \emph{ICML}, 2020.

\bibitem[Rissanen(1996)]{rissanen1996fisher}
Rissanen, J.~J.
\newblock Fisher information and stochastic complexity.
\newblock \emph{IEEE transactions on information theory}, 42\penalty0
  (1):\penalty0 40--47, 1996.

\bibitem[Ruder(2016)]{ruder2016overview}
Ruder, S.
\newblock An overview of gradient descent optimization algorithms.
\newblock \emph{arXiv preprint arXiv:1609.04747}, 2016.

\bibitem[Schervish(2012)]{schervish2012theory}
Schervish, M.~J.
\newblock \emph{Theory of statistics}.
\newblock Springer Science \& Business Media, 2012.

\bibitem[Sch{\"o}lkopf et~al.(2002)Sch{\"o}lkopf, Smola, Bach,
  et~al.]{scholkopf2002learning}
Sch{\"o}lkopf, B., Smola, A.~J., Bach, F., et~al.
\newblock \emph{Learning with kernels: support vector machines, regularization,
  optimization, and beyond}.
\newblock MIT press, 2002.

\bibitem[Shen et~al.(2020)Shen, Wang, Ribeiro, and Hassani]{shen2020sinkhorn}
Shen, Z., Wang, Z., Ribeiro, A., and Hassani, H.
\newblock Sinkhorn barycenter via functional gradient descent.
\newblock In \emph{NeurIPS}, 2020.

\bibitem[Singer(1974)]{singer1974theory}
Singer, I.
\newblock \emph{The theory of best approximation and functional analysis}.
\newblock SIAM, 1974.

\bibitem[Singla et~al.(2013)Singla, Bogunovic, Bart{\'o}k, Karbasi, and
  Krause]{singla2013actively}
Singla, A., Bogunovic, I., Bart{\'o}k, G., Karbasi, A., and Krause, A.
\newblock On actively teaching the crowd to classify.
\newblock In \emph{NIPS Workshop on Data Driven Education}, 2013.

\bibitem[Singla et~al.(2014)Singla, Bogunovic, Bart{\'o}k, Karbasi, and
  Krause]{singla2014near}
Singla, A., Bogunovic, I., Bart{\'o}k, G., Karbasi, A., and Krause, A.
\newblock Near-optimally teaching the crowd to classify.
\newblock In \emph{ICML}, 2014.

\bibitem[Smanski et~al.(2014)Smanski, Bhatia, Zhao, Park, BA~Woodruff,
  Giannoukos, Ciulla, Busby, Calderon, Nicol, et~al.]{smanski2014functional}
Smanski, M.~J., Bhatia, S., Zhao, D., Park, Y., BA~Woodruff, L., Giannoukos,
  G., Ciulla, D., Busby, M., Calderon, J., Nicol, R., et~al.
\newblock Functional optimization of gene clusters by combinatorial design and
  assembly.
\newblock \emph{Nature biotechnology}, 32\penalty0 (12):\penalty0 1241--1249,
  2014.

\bibitem[Steinwart \& Christmann(2008)Steinwart and
  Christmann]{steinwart2008support}
Steinwart, I. and Christmann, A.
\newblock \emph{Support vector machines}.
\newblock Springer Science \& Business Media, 2008.

\bibitem[Tabibian et~al.(2019)Tabibian, Upadhyay, De, Zarezade, Sch{\"o}lkopf,
  and Gomez-Rodriguez]{tabibian2019enhancing}
Tabibian, B., Upadhyay, U., De, A., Zarezade, A., Sch{\"o}lkopf, B., and
  Gomez-Rodriguez, M.
\newblock Enhancing human learning via spaced repetition optimization.
\newblock \emph{Proceedings of the National Academy of Sciences}, 116\penalty0
  (10):\penalty0 3988--3993, 2019.

\bibitem[Vajda(1989)]{vajda1989theory}
Vajda, I.
\newblock \emph{Theory of statistical inference and information}.
\newblock Springer, 1989.

\bibitem[Vajda(2002)]{vajda2002convergence}
Vajda, I.
\newblock On convergence of information contained in quantized observations.
\newblock \emph{IEEE Transactions on Information Theory}, 48\penalty0
  (8):\penalty0 2163--2172, 2002.

\bibitem[Wang \& Vasconcelos(2021)Wang and Vasconcelos]{wang2021machine}
Wang, P. and Vasconcelos, N.
\newblock A machine teaching framework for scalable recognition.
\newblock In \emph{ICCV}, 2021.

\bibitem[Wang et~al.(2021)Wang, Nagrecha, and Vasconcelos]{wang2021gradient}
Wang, P., Nagrecha, K., and Vasconcelos, N.
\newblock Gradient-based algorithms for machine teaching.
\newblock In \emph{CVPR}, 2021.

\bibitem[Xu et~al.(2021)Xu, Chen, Li, Liu, Song, Lin, and
  Shrivastava]{xu2021locality}
Xu, Z., Chen, B., Li, C., Liu, W., Song, L., Lin, Y., and Shrivastava, A.
\newblock Locality sensitive teaching.
\newblock In \emph{NeurIPS}, 2021.

\bibitem[Ye et~al.(2017)Ye, Wu, Wang, and Li]{ye2017fast}
Ye, J., Wu, P., Wang, J.~Z., and Li, J.
\newblock Fast discrete distribution clustering using wasserstein barycenter
  with sparse support.
\newblock \emph{IEEE Transactions on Signal Processing}, 65\penalty0
  (9):\penalty0 2317--2332, 2017.

\bibitem[Zhang et~al.(2020{\natexlab{a}})Zhang, Ye, Gong, Zhu, and
  Liu]{zhang2020black}
Zhang, D., Ye, M., Gong, C., Zhu, Z., and Liu, Q.
\newblock Black-box certification with randomized smoothing: A functional
  optimization based framework.
\newblock In \emph{NeurIPS}, 2020{\natexlab{a}}.

\bibitem[Zhang et~al.(2020{\natexlab{b}})Zhang, Bharti, Ma, Singla, and
  Zhu]{zhang2020sample}
Zhang, X., Bharti, S.~K., Ma, Y., Singla, A., and Zhu, X.
\newblock The sample complexity of teaching-by-reinforcement on q-learning.
\newblock \emph{arXiv preprint arXiv:2006.09324}, 2020{\natexlab{b}}.

\bibitem[Zhou et~al.(2018)Zhou, Nelakurthi, and He]{zhou2018unlearn}
Zhou, Y., Nelakurthi, A.~R., and He, J.
\newblock Unlearn what you have learned: Adaptive crowd teaching with
  exponentially decayed memory learners.
\newblock In \emph{SIGKDD}, 2018.

\bibitem[Zhou et~al.(2020)Zhou, Nelakurthi, Maciejewski, Fan, and
  He]{zhou2020crowd}
Zhou, Y., Nelakurthi, A.~R., Maciejewski, R., Fan, W., and He, J.
\newblock Crowd teaching with imperfect labels.
\newblock In \emph{WWW}, 2020.

\bibitem[Zhu(2013)]{zhu2013machine}
Zhu, X.
\newblock Machine teaching for bayesian learners in the exponential family.
\newblock \emph{arXiv preprint arXiv:1306.4947}, 2013.

\bibitem[Zhu(2015)]{zhu2015machine}
Zhu, X.
\newblock Machine teaching: An inverse problem to machine learning and an
  approach toward optimal education.
\newblock In \emph{AAAI}, 2015.

\bibitem[Zhu et~al.(2017)Zhu, Liu, and Lopes]{zhu2017no}
Zhu, X., Liu, J., and Lopes, M.
\newblock No learner left behind: On the complexity of teaching multiple
  learners simultaneously.
\newblock In \emph{IJCAI}, 2017.

\bibitem[Zhu et~al.(2018)Zhu, Singla, Zilles, and Rafferty]{zhu2018overview}
Zhu, X., Singla, A., Zilles, S., and Rafferty, A.~N.
\newblock An overview of machine teaching.
\newblock \emph{arXiv preprint arXiv:1801.05927}, 2018.

\bibitem[Zoppoli et~al.(2002)Zoppoli, Sanguineti, and
  Parisini]{zoppoli2002approximating}
Zoppoli, R., Sanguineti, M., and Parisini, T.
\newblock Approximating networks and extended ritz method for the solution of
  functional optimization problems.
\newblock \emph{Journal of Optimization Theory and Applications}, 112\penalty0
  (2):\penalty0 403--440, 2002.

\end{thebibliography}
\bibliographystyle{icml2023}

\clearpage
\newpage

\appendix
\onecolumn

\newpage

\begin{appendix}
	
	\thispagestyle{plain}
	\begin{center}
		{\Large \bf Appendix}
	\end{center}
	
\end{appendix}

\section{Additional Discussions}
\textbf{Broader Impact} Machine teaching has been applied in crowd sourcing, computer vision and cyber security -- domains with significant societal impacts. This work focuses on theoretical analysis of iterative machine teaching and generalizes parameterized iterative machine teaching to nonparametric scenarios, which is to generalize model space from a finite dimensional one to an infinite dimensional one. This provides possibility of extending parameterized applications to nonparametric cases. Thus, while the contributions of this work are mainly theoretical, there are potential positive impacts in the community of machine teaching and society.

\textbf{Parametric teaching settings}\label{parat}
One can rewrite formulations in Section~\ref{ts} into parameterized version via replacing $f$ by $w$ \cite{liu2017iterative,liu2018towards,zhu2018overview} as parametric IMT operates in the finite dimensional parameter space. Specifically, the bilevel optimization can be formulated as 
\begin{eqnarray}
	\mathcal{D}^*=\underset{\mathcal{D}\in\mathbb{D}}{\arg\min}\, \mathcal{M}(\hat{w},w^*)+\lambda\cdot \text{len}(\mathcal{D})\quad\text{s.t.}\,\hat{w}=\mathcal{A}(\mathcal{D}),
\end{eqnarray} where notations have same meanings as Eq.~\ref{eq1}. Empirical risk minimization $\mathcal{A}(\mathcal{D})$ is as follows
\begin{eqnarray}
	\hat{w^*}=\underset {w}{\arg\min}\,\mathbb{E}_{(\bm{x},y)}\left\{\mathcal{L}(\langle w, \bm{x}\rangle,y)\right\}.
\end{eqnarray}
Besides, parameter $w$ is updated as 
\begin{equation}
	w^{t+1}\gets w^t-\eta^t\mathcal{G}(\mathcal{L};w^t;(\bm{x}^t,y^t)).
\end{equation}

\textbf{Nonparametric Fisher information for convex loss function in Section~\ref{ftca}} \label{nlfi}
$\quad$ Fisher information \cite{lehmann2006theory} is a fundamental quantity in statistics and information theory \cite{vajda1989theory}. It measures the information carried by data about an unknown parameter $\theta$. Let \begin{equation}\label{fi}
	\mathcal{I}(\theta) = \mathbb{E}_{\bm{x}\sim\mathbb{P}(\bm{x})}\left\{\left(\nabla_\theta \log\ell(\bm{x};\theta)\right)^2\right\}
\end{equation} be Fisher information. It can be written \cite{vajda2002convergence} as 
\begin{equation}\label{phifi}
	\mathcal{I}_\phi(\theta) = \mathbb{E}_{\bm{x}\sim\mathbb{P}(\bm{x})}\left\{\phi\left(\nabla_\theta \log\ell(\bm{x};\theta)\right)\right\},
\end{equation}
where $\phi(\cdot) = (\cdot)^2$. There are many works \cite{noguchi1994invariant, vajda2002convergence, lutwak2012extensions, lv2017general} on generalized Fisher information in terms of explicit form of $\phi(\cdot)$. For example, \citealp{kallenberg1985number} considers $\phi(\cdot)=(\cdot)^{4/3}$ and connects it with Pearson goodness of fit test. Besides, let $\phi(\cdot)=-\log(\cdot)$, Eq.~\ref{fi} is the information divergence \cite{vajda2002convergence}.

From another generalized perspective, Eq.~\ref{fi} can be rewritten in another way as 
\begin{eqnarray}
	\mathcal{I}(\theta)_{\varphi;\vartheta} = \mathbb{E}_{\bm{x}\sim\mathbb{P}(\bm{x})}\left\{\left(\nabla_\vartheta \varphi(\bm{x};\vartheta)\right)^2\right\},
\end{eqnarray} where $\varphi(\cdot) = \log\ell(\cdot)$ and $\vartheta=\theta$. Meanwhile, concerned with arithmetic mean instead of point evaluation of $\left(\nabla_f \mathcal{L}(f)\right)^2$, $\mathbb{S}_{\mathcal{L}}(f;\bm{x})=\left|E_{\bm{x}}\nabla_f \mathcal{L}(f)\right|^2$ introduced in Definition~\ref{disc} can be written as 
\begin{equation}
	\mathbb{E}_{\bm{x}\sim\mathbb{P}(\bm{x})}\left\{\left(\nabla_f \mathcal{L}(\bm{x};f)\right)^2\right\},
\end{equation}
which can be viewed as a nonparametric Fisher information for convex loss function. Therefore, extending $\varphi(\cdot)$ from the natural logarithm of the likelihood function, $\log\ell(\cdot)$ to the convex loss function $\mathcal{L}(\cdot)$ and extending $\vartheta$ from unknown parameter $\theta$ to general mapping $f$, nonparametric Fisher information for convex loss function can be viewed as a kind of generalized Fisher information.

\section{Detailed Proofs}

We recommend the literature \cite{ gelfand2000calculus, coleman2012calculus} for further reading on functional calculus. 

\textbf{Proof of Lemma~\ref{ef}}\label{pef}
$\quad$Let define a function $q$ by adding a small perturbation $\epsilon g$ ($\epsilon\in\mathbb{R}, g\in\mathcal{H}$) to $f\in\mathcal{H}$, $q=f+\epsilon g$. $q\in\mathcal{H}$ since RKHS is closed under addition and scalar multiplication. Therefore, for a evaluation functional $E_{\bm{x}}[f]=f(\bm{x}):\mathcal{H}\mapsto\mathbb{R}$, we can evaluate $q$ at $\bm{x}$ as 
\begin{eqnarray}
	\label{eqpef}
	E_{\bm{x}}[q]&=&E_{\bm{x}}[f+\epsilon g]\nonumber\\
	&=&E_{\bm{x}}[f]+\epsilon E_{\bm{x}}[g]+0\nonumber\\
	&=&E_{\bm{x}}[f]+\epsilon \langle K(\bm{x},\cdot),g\rangle_\mathcal{H}+0
\end{eqnarray}
Recall implicit definition of Fréchet derivative in RKHS (see Definition~\ref{defn1}) $E_{\bm{x}}[f+\epsilon g]=E_{\bm{x}}[f]+\epsilon \langle \nabla_f E_{\bm{x}}[f],g\rangle_\mathcal{H}+\mathcal{O}(\epsilon^2)$, it follows from Eq.~\ref{eqpef} that we have the gradient of a evaluation functional $\nabla_f E_{\bm{x}}[f] = K_{\bm{x}}$.

$\blacksquare$

\textbf{Proof of Theorem~\ref{optthm}}\label{poptthm}
$\quad$Concisely, we omit superscript $t$ for the time being and rewrite Eq.~\ref{optx} as 
\begin{equation}
	({\bm{x}}^*,{y}^*) = \underset {\bm{x}\in\mathcal{X},y\in\mathcal{Y}}{\arg\min}\left\|f-\eta\cdot \mathcal{G}-f^*\right\|^2_\mathcal{H}.
\end{equation}
Obviously, it is trivial to derive that  $\forall (\bm{x},y), \bm{x}\in\mathcal{X},y\in\mathcal{Y},$
\begin{eqnarray}\label{poptthmeq1}
	\left\|f-\eta \mathcal{G}(\mathcal{L};f;(\bm{x}^*,y^*))-f^*\right\|^2_\mathcal{H} \leq \nonumber\\
	\left\|f-\eta \mathcal{G}(\mathcal{L};f;(\bm{x},y))-f^*\right\|^2_\mathcal{H}.
\end{eqnarray}
Out of succinctness, we denote ${\mathcal{G}^t}^*\coloneqq\mathcal{G}(\mathcal{L};f^t;({\bm{x}^t}^*,{y^t}^*))$ and $\mathcal{G}^t\coloneqq\mathcal{G}(\mathcal{L};f^t;(\bm{x}^t,y^t))$.
For l.h.s. of expression~\ref{poptthmeq1}, we can expand it as
\begin{eqnarray}
	&&\left\|f-\eta \mathcal{G}(\mathcal{L};f;(\bm{x}^*,y^*))-f^*\right\|^2_\mathcal{H}\nonumber\\
	&=&\left\|f-f^*\right\|^2_\mathcal{H}+\eta^2\left\|\mathcal{G}^*\right\|_\mathcal{H}^2-\eta \left\langle\mathcal{G}^*,f-f^*\right\rangle_\mathcal{H}.
\end{eqnarray}
Similarly, we can also expand r.h.s. of expression~\ref{poptthmeq1} as
\begin{eqnarray}
	&&\left\|f-\eta \mathcal{G}(\mathcal{L};f;(\bm{x},y))-f^*\right\|^2_\mathcal{H}\nonumber\\
	&=&\left\|f-f^*\right\|^2_\mathcal{H}+\eta^2\left\|\mathcal{G}\right\|_\mathcal{H}^2-\eta \left\langle\mathcal{G},f-f^*\right\rangle_\mathcal{H}.
\end{eqnarray}
Combining expansion of expression~\ref{poptthmeq1} together, we have 
\begin{eqnarray}
	&&\left\|f-f^*\right\|^2_\mathcal{H}+\eta^2\left\|\mathcal{G}^*\right\|_\mathcal{H}^2-\eta \left\langle\mathcal{G}^*,f-f^*\right\rangle_\mathcal{H}\nonumber\\
	&\leq&\left\|f-f^*\right\|^2_\mathcal{H}+\eta^2\left\|\mathcal{G}\right\|_\mathcal{H}^2-\eta \left\langle\mathcal{G},f-f^*\right\rangle_\mathcal{H}.
\end{eqnarray}
After rearranging, we can obtain
\begin{eqnarray}
	\langle \mathcal{G}^*-\mathcal{G},f-f^*\rangle_\mathcal{H}\geq\eta/2(\left\|\mathcal{G}^*\right\|^2_\mathcal{H}-\left\|\mathcal{G}\right\|^2_\mathcal{H})\geq0.
\end{eqnarray}
$\blacksquare$

\textbf{Proof of Lemma~\ref{sdpft}}\label{psdpft}
$\quad$  Recall the definition of Fréchet derivative in Definition~\ref{defn1}. It follows from the convexity of $\mathcal{L}$ that we have
\begin{eqnarray}\label{psdpfteq1}
	\mathcal{L}(f^{t+1})-\mathcal{L}(f^t)\leq\langle f^{t+1}-f^t,\left.\nabla_f\mathcal{L}(f)\right|_{f=f^{t+1}}\rangle_\mathcal{H}.
\end{eqnarray}
Based on optimization algorithm in Eq.~\ref{opta}, the right term of Eq.~\ref{psdpfteq1} can be expressed as
\begin{eqnarray}
	\langle f^{t+1}-f^t,\left.\nabla_f\mathcal{L}(f)\right|_{f=f^{t+1}}\rangle_\mathcal{H}
	=\langle -\eta^t\mathcal{G}^t,\left.\nabla_f\mathcal{L}(f)\right|_{f=f^{t+1}}\rangle_\mathcal{H}\nonumber.
\end{eqnarray}
Substituting $\mathcal{G}^t= \left.\frac{\partial\mathcal{L}}{\partial f}\right|_{f^t(\bm{x}^t),y^t} \cdot K (\bm{x}^t,\cdot)$ in and removing constants out of inner product operation, it yields
\begin{eqnarray} \label{psdpfteq2}
	&&\langle -\eta^t\mathcal{G}^t,\left.\nabla_f\mathcal{L}(f)\right|_{f=f^{t+1}}\rangle_\mathcal{H}\nonumber\\
	&=&-\eta^t\left\langle \left.\frac{\partial\mathcal{L}}{\partial f}\right|_{f^t(\bm{x}^t),y^t} \cdot K (\bm{x}^t,\cdot),\nabla_f\mathcal{L}(f)|_{f=f^{t+1}}\right\rangle_\mathcal{H}\nonumber\\
	&=&-\eta^t\left.\frac{\partial\mathcal{L}}{\partial f}\right|_{f^t(\bm{x}^t),y^t}\left\langle K (\bm{x}^t,\cdot),\nabla_f\mathcal{L}(f)|_{f=f^{t+1}}\right\rangle_\mathcal{H}.
\end{eqnarray}
Recall the definition of the evaluation functional in RKHS in Definition~\ref{efl}
\begin{equation}
	E_{\bm{x}}[f]=\langle f, K_{\bm{x}}(\cdot)\rangle_\mathcal{H}
\end{equation}
and the fact $y=f^*(\bm{x})=E_{\bm{x}}[f^*]$, we can rewrite the last term in Eq.~\ref{psdpfteq2} as
\begin{eqnarray}\label{psdpfteq3}
	&&-\eta^t\left.\frac{\partial\mathcal{L}}{\partial f}\right|_{f^t(\bm{x}^t),y^t}\left\langle K (\bm{x}^t,\cdot),\nabla_f\mathcal{L}(f)|_{f=f^{t+1}}\right\rangle_\mathcal{H}\nonumber\\
	&=&	-\eta^t\left.\frac{\partial\mathcal{L}}{\partial f}\right|_{f^t(\bm{x}^t),y^t}\times E_{\bm{x}^t}\left[\left.\nabla_f\mathcal{L}(f)\right|_{f=f^{t+1}}\right]\nonumber\\
	&=&-\eta^t E_{\bm{x}^t}\left[\left.\nabla_f\mathcal{L}(f)\right|_{f=f^{t}}\right]\times E_{\bm{x}^t}\left[\left.\nabla_f\mathcal{L}(f)\right|_{f=f^{t+1}}\right].
\end{eqnarray}
For succinctness, denote $\xi^{t}\coloneqq E_{\bm{x}^t}\left[\left.\nabla_f\mathcal{L}(f)\right|_{f=f^{t}}\right]$ and $\xi^{t+1}\coloneqq E_{\bm{x}^t}\left[\left.\nabla_f\mathcal{L}(f)\right|_{f=f^{t+1}}\right]$, then Eq.~\ref{itd} can be tersely expressed
$\mathbb{S}_{\mathcal{L}}(f^t;{\bm{x}^t})=\left|E_{\bm{x}^t}\left[\left.\nabla_f\mathcal{L}(f)\right|_{f=f^{t}}\right]\right|^2=(\xi^t)^2$, and we thus can rewrite Eq.~\ref{psdpfteq3} as follows:
\begin{eqnarray}
	&&-\eta^t E_{\bm{x}^t}\left[\left.\nabla_f\mathcal{L}(f)\right|_{f=f^{t}}\right]\times E_{\bm{x}^t}\left[\left.\nabla_f\mathcal{L}(f)\right|_{f=f^{t+1}}\right]\nonumber\\
	&=& -\eta^t\xi^t\times\xi^{t+1}\nonumber\\
	&=&	-\eta^t \xi^t\times\left(\xi^t+\xi^{t+1}-\xi^t\right)\nonumber\\
	&=&	-\eta^t \mathbb{S}_{\mathcal{L}}(f^t;{\bm{x}^t})-\eta^t\xi^t\times\left(\xi^{t+1}-\xi^t\right)\nonumber\\
	&=&	-\eta^t \mathbb{S}_{\mathcal{L}}(f^t;{\bm{x}^t})+\eta^t\left(\xi^{t+1}-\xi^t-\xi^{t+1}\right)\times\left(\xi^{t+1}-\xi^t\right)\nonumber\\
	&=&	-\eta^t \mathbb{S}_{\mathcal{L}}(f^t;{\bm{x}^t})+\eta^t\left(\xi^{t+1}-\xi^t\right)^2-\eta^t\xi^{t+1}\left(\xi^{t+1}-\xi^t\right)\nonumber\\
	&=&	-\eta^t \mathbb{S}_{\mathcal{L}}(f^t;{\bm{x}^t})+\eta^t\left(\xi^{t+1}-\xi^t\right)^2-\eta^t\left(\xi^{t+1}-1/2\xi^t\right)^2+1/4\eta^t\mathbb{S}_{\mathcal{L}}(f^t;{\bm{x}^t})\nonumber\\
	&=&	-3/4\eta^t \mathbb{S}_{\mathcal{L}}(f^t;{\bm{x}^t})+\eta^t\left(\xi^{t+1}-\xi^t\right)^2-\eta^t\left(\xi^{t+1}-1/2\xi^t\right)^2\nonumber\\
	&\leq&	-3/4\eta^t \mathbb{S}_{\mathcal{L}}\left(f^t;{\bm{x}^t}\right)+\eta^t\left(\xi^{t+1}-\xi^t\right)^2.
\end{eqnarray}
Substituting the concrete expression of  $\xi^{t}= E_{\bm{x}^t}\left[\left.\nabla_f\mathcal{L}(f)\right|_{f=f^{t}}\right]$ and $\xi^{t+1}= E_{\bm{x}^t}\left[\left.\nabla_f\mathcal{L}(f)\right|_{f=f^{t+1}}\right]$ in, it follows from linearity of evaluation functional that
\begin{eqnarray}
	&&-3/4\eta^t \mathbb{S}_{\mathcal{L}}(f^t;{\bm{x}^t})+\eta^t\left(\xi^{t+1}-\xi^t\right)^2\nonumber\\
	&=&	-3/4\eta^t \mathbb{S}_{\mathcal{L}}(f^t;{\bm{x}^t})+\eta^t\left(E_{\bm{x}^t}\left[\left.\nabla_f\mathcal{L}(f)\right|_{f=f^{t+1}}\right]-E_{\bm{x}^t}\left[\left.\nabla_f\mathcal{L}(f)\right|_{f=f^{t}}\right]\right)^2\nonumber\\
	&=&	-3/4\eta^t \mathbb{S}_{\mathcal{L}}(f^t;{\bm{x}^t})+\eta^t\left(E_{\bm{x}^t}\left[\left.\nabla_f\mathcal{L}(f)\right|_{f=f^{t+1}}-\left.\nabla_f\mathcal{L}(f)\right|_{f=f^{t}}\right]\right)^2.
\end{eqnarray}
Under L-Lipschitz smooth Assumption~\ref{lls} and bounded kernel function Assumption~\ref{bkf}, we have
\begin{eqnarray}
	&&-3/4\eta^t \mathbb{S}_{\mathcal{L}}(f^t;{\bm{x}^t})+\eta^t\left(E_{\bm{x}^t}\left[\left.\nabla_f\mathcal{L}(f)\right|_{f=f^{t+1}}-\left.\nabla_f\mathcal{L}(f)\right|_{f=f^{t}}\right]\right)^2\nonumber\\
	&\leq&	-3/4\eta^t \mathbb{S}_{\mathcal{L}}(f^t;{\bm{x}^t})+\eta^t\left(L_\mathcal{L}\cdot E_{\bm{x}^t}\left[\left|f^{t+1}-f^t\right|\right]\right)^2\nonumber\\
	&=&	-3/4\eta^t \mathbb{S}_{\mathcal{L}}(f^t;{\bm{x}^t})+\eta^t\left(L_\mathcal{L}\eta^t\xi^t E_{\bm{x}^t}\left[K_{\bm{x}^t}\right]\right)^2\nonumber\\
	&=&	-3/4\eta^t \mathbb{S}_{\mathcal{L}}(f^t;{\bm{x}^t})+L_\mathcal{L}^2\left(\eta^t\right)^3\mathbb{S}_{\mathcal{L}}(f^t;{\bm{x}^t})K^2\left(\bm{x}^t,\bm{x}^t\right)\nonumber\\
	&\leq&	-3/4\eta^t \mathbb{S}_{\mathcal{L}}(f^t;{\bm{x}^t})+L_\mathcal{L}^2\left(\eta^t\right)^3\mathbb{S}_{\mathcal{L}}(f^t;{\bm{x}^t})(M_K)^2\nonumber\\
	&=&	-\eta^t\left(3/4-L_\mathcal{L}^2(\eta^t)^2 (M_K)^2\right) \mathbb{S}_{\mathcal{L}}(f^t;{\bm{x}^t}).
\end{eqnarray}

Consequently, we obtain
\begin{eqnarray}
	\mathcal{L}(f^{t+1})-\mathcal{L}(f^t)\leq-\eta^t\left(3/4-L_\mathcal{L}^2(\eta^t)^2 M^2_K\right) \mathbb{S}_{\mathcal{L}}(f^t;{\bm{x}^t}),
\end{eqnarray}
and hence $\mathcal{L}(f^{t+1})-\mathcal{L}(f^t)\leq-\eta^t/2\cdot\mathbb{S}_{\mathcal{L}}(f^t;{\bm{x}^t})$ if $\eta^t\leq \frac{1}{2L_\mathcal{L}\cdot M_K}$

$\blacksquare$

\textbf{Proof of Theorem~\ref{cpft}}\label{pcpft}
$\quad$ Recall Lemma~\ref{sdpft}, when $\eta^t\leq \frac{1}{2L_\mathcal{L}\cdot M_K}$,  
\begin{equation}
	\mathcal{L}(f^{t+1})-\mathcal{L}(f^t)\leq-\eta^t/2\cdot\mathbb{S}_{\mathcal{L}}(f^t;{\bm{x}^t})
\end{equation}
Rearranging above, we have:
\begin{eqnarray}
	\frac{2\left(\mathcal{L}(f^t)-\mathcal{L}(f^{t+1})\right)}{\eta^t}\geq\mathbb{S}_{\mathcal{L}}(f^t;{\bm{x}^t}).
\end{eqnarray}
Equivalently, $\frac{2\left(\mathcal{L}(f^j)-\mathcal{L}(f^{j+1})\right)}{\eta^j}\geq\mathbb{S}_{\mathcal{L}}(f^j;{\bm{x}^j})$.
Consequently, plugging $j=0,1\dots,t-1$ in it and summing them up, we hence have 
\begin{eqnarray}\label{psdpfteq4}
	\sum_{j=0}^{t-1}\mathbb{S}_{\mathcal{L}}(f^j;{\bm{x}^j})	\leq 2\sum_{j=0}^{t-1}\frac{\mathcal{L}(f^j)-\mathcal{L}(f^{j+1})}{\eta^j}\leq \frac{2}{\tilde{\eta}}\sum_{j=0}^{t-1}\left(\mathcal{L}(f^j)-\mathcal{L}(f^{j+1})\right),
\end{eqnarray}
where $\tilde{\eta}=\underset{j}{\min}\,\eta^j>0$. Expanding the r.h.s. term in Eq.~\ref{psdpfteq4} yields
\begin{eqnarray}\label{psdpfteq5}
	\frac{2}{\tilde{\eta}}\sum_{j=0}^{t-1}\left(\mathcal{L}(f^j)-\mathcal{L}(f^{j+1})\right)=\frac{2}{\tilde{\eta}}\left(\mathcal{L}(f^0)-\mathcal{L}(f^{t})\right)\leq\frac{2}{\tilde{\eta}}\mathcal{L}(f^0).
\end{eqnarray}
In terms of the l.h.s. term in Eq.~\ref{psdpfteq4}, we must have
\begin{eqnarray}\label{psdpfteq6}
	\sum_{j=0}^{t-1}\mathbb{S}_{\mathcal{L}}(f^j;{\bm{x}^j})\geq t\cdot\min_j\mathbb{S}_{\mathcal{L}}(f^j;{\bm{x}^j}).
\end{eqnarray}
Combining expression~\ref{psdpfteq5} and \ref{psdpfteq6}, we thus have
\begin{eqnarray}
	t\cdot\min_t\mathbb{S}_{\mathcal{L}}(f^t;{\bm{x}^t})\leq\sum_{j=0}^{t-1}\mathbb{S}_{\mathcal{L}}(f^j;{\bm{x}^j})\leq\frac{2}{\tilde{\eta}}\mathcal{L}(f^0),
\end{eqnarray}
from which, we can derive
\begin{eqnarray}
	\min_t\mathbb{S}_{\mathcal{L}}(f^t;{\bm{x}^t})\leq2\mathcal{L}(f^0)/\left(\tilde{\eta}t\right).
\end{eqnarray}
$\blacksquare$

It suggests that learners could also conduct its stationary state as: In each round, check if $\mathbb{S}_{\mathcal{L}}(f^t;{\bm{x}^t})\leq\epsilon$. If it holds, then they have already reached the $\epsilon$ approximating and they can send a terminated signal to teachers; otherwise teachers proceed. The termination occurs within $\mathcal{O}(2\mathcal{L}(f^0)/\left(\tilde{\eta}\epsilon\right))$ loops.

\textbf{Proof of Lemma~\ref{sdaft}}\label{psdaft}
$\quad$ Recall practical Greedy Functional Teaching in Eq.~\ref{aft1prac}
\begin{equation}
	\left({\bm{x}^t}^*=\underset{\bm{x}^t\in\mathcal{X}}{\arg\max}\left| E_{\bm{x}^t}\left[\left.\nabla_f\mathcal{L}(f)\right|_{f=f^{t}}\right] \right|,y^*=E_{{\bm{x}^t}^*}\left[f^*\right]\right).
\end{equation}
Obviously, it is trivial to see that $\forall \bm{x}^t\in\mathcal{X}$,
\begin{eqnarray}
	\left|E_{{\bm{x}^t}^*}\left[\left.\nabla_f\mathcal{L}(f)\right|_{f=f^{t}}\right]\right|^2\geq\left|E_{\bm{x}^t}\left[\left.\nabla_f\mathcal{L}(f)\right|_{f=f^{t}}\right]\right|^2.
\end{eqnarray}
Analogous to the Proof of Lemma~\ref{sdpft} in \ref{psdpft}, we can derive
\begin{eqnarray}
	\mathcal{L}(f^{t+1})-\mathcal{L}(f^t)&\leq&-\eta^t/2\cdot\mathbb{S}_{\mathcal{L}}(f^t;{\bm{x}^t}^*),
\end{eqnarray}
if $0<\eta^t\leq \frac{1}{2L_\mathcal{L}\cdot M_K}$. Consequently, we have 
\begin{eqnarray}
	\mathcal{L}(f^{t+1})-\mathcal{L}(f^t)\leq-\eta^t/2\cdot\mathbb{S}_{\mathcal{L}}(f^t;{\bm{x}^t}^*)\leq-\eta^t/2\cdot\mathbb{S}_{\mathcal{L}}(f^t;{\bm{x}^t}).
\end{eqnarray}
$\blacksquare$

\textbf{Proof of Theorem~\ref{caft}}\label{pcaft}
$\quad$ Recall the result of Lemma~\ref{sdpft}, when $0<\eta^t\leq \frac{1}{2L_\mathcal{L}\cdot M_K}$
\begin{equation}
	\mathbb{S}_{\mathcal{L}}(f^t;{\bm{x}^t})\leq\frac{2\left(\mathcal{L}(f^t)-\mathcal{L}(f^{t+1})\right)}{\eta^t}.
\end{equation}
Before converging to the stationary state, $\mathbb{S}_{\mathcal{L}}(f^t;{\bm{x}^t}^*)>0$. Therefore, we can express it as
\begin{equation}
	\mathbb{S}_{\mathcal{L}}(f^t;{\bm{x}^t}^*)\cdot\frac{\mathbb{S}_{\mathcal{L}}(f^t;{\bm{x}^t})}{\mathbb{S}_{\mathcal{L}}(f^t;{\bm{x}^t}^*)}\leq\frac{2\left(\mathcal{L}(f^t)-\mathcal{L}(f^{t+1})\right)}{\eta^t}.
\end{equation}

For succinctness, denote $\gamma^t \coloneqq 
\frac{\mathbb{S}_{\mathcal{L}}(f^t;{\bm{x}^t})}{\mathbb{S}_{\mathcal{L}}(f^t;{\bm{x}^t}^*)}$, namely greedy ratio, we have
\begin{eqnarray}
	\gamma^t\cdot\mathbb{S}_{\mathcal{L}}(f^t;{\bm{x}^t}^*)\leq \frac{2\left(\mathcal{L}(f^t)-\mathcal{L}(f^{t+1})\right)}{\eta^t}.
\end{eqnarray}
Different to expression~\ref{psdpfteq6}, we have 
\begin{eqnarray}
	\sum_{j=0}^{t-1}\gamma^j\cdot\mathbb{S}_{\mathcal{L}}(f^j;{\bm{x}^j}^*)\geq \min_j\mathbb{S}_{\mathcal{L}}(f^j;{\bm{x}^j}^*)\cdot\sum_{j=0}^{t-1}\gamma^j.
\end{eqnarray}
Since ${\bm{x}^t}^*=\underset{\bm{x}^t\in\mathcal{X}}{\arg\max}\left\|\left.\frac{\partial\mathcal{L}}{\partial f}\right|_{f^t(\bm{x}^t),y^*}K \left(\bm{x}^t,\cdot\right)\right\|_\mathcal{H}$, we derive $\left|E_{\bm{x}^t}\nabla_f \mathcal{L}(f^t,f^*)\right|^2 \leq \left|E_{{\bm{x}^t}^*}\nabla_f \mathcal{L}(f^t,f^*)\right|^2$. Therefore, $\gamma^t = 
\frac{\mathbb{S}_{\mathcal{L}}(f^t;{\bm{x}^t})}{\mathbb{S}_{\mathcal{L}}(f^t;{\bm{x}^t}^*)}=\frac{\left|E_{\bm{x}^t}\nabla_f \mathcal{L}(f^t,f^*)\right|^2}{\left|E_{{\bm{x}^t}^*}\nabla_f \mathcal{L}(f^t,f^*)\right|^2}\in(0,1]$ and we have $\min_j\mathbb{S}_{\mathcal{L}}(f^j;{\bm{x}^j}^*)\cdot\sum_{j=0}^{t-1}\gamma^j\leq t\cdot\min_j\mathbb{S}_{\mathcal{L}}(f^j;{\bm{x}^j}^*)$. 
Besides, similar to expression~\ref{psdpfteq5}, we have 
\begin{eqnarray}
	\frac{2}{\tilde{\eta}}\sum_{j=0}^{t-1}\left(\mathcal{L}(f^j)-\mathcal{L}(f^{j+1})\right)=\frac{2}{\tilde{\eta}}\left(\mathcal{L}(f^0)-\mathcal{L}(f^{t})\right)\leq\frac{2}{\tilde{\eta}}\mathcal{L}(f^0),
\end{eqnarray}
where  $\tilde{\eta}=\underset{j}{\min}\,\eta^j>0$. To sum up, we have
\begin{eqnarray}
	\min_j\mathbb{S}_{\mathcal{L}}(f^j;{\bm{x}^j}^*)\cdot\sum_{j=0}^{t-1}\gamma^j\leq\frac{2}{\tilde{\eta}}\mathcal{L}(f^0),
\end{eqnarray}
For succinctness, denote $\psi(t)\coloneqq\sum_{j=0}^{t-1}\gamma^j$. Note that $\underset{{t\to \infty}}{\lim}\gamma^t\to1\Rightarrow \underset{{t\to \infty}}{\lim}\psi(t)=\underset{{t\to \infty}}{\lim}\sum_{j=0}^{t-1}\gamma^j\to\infty$. Rearranging, we obtain
\begin{eqnarray}
	\min_j\mathbb{S}_{\mathcal{L}}(f^j;{\bm{x}^j}^*)\leq\frac{2}{\tilde{\eta}\psi(t)}\mathcal{L}(f^0).
\end{eqnarray}
Since $\psi(t)=\sum_{j=0}^{t-1}\gamma^j\leq t \cdot \max_j \gamma^j$ and $\gamma^j\in (0,1]$, we have $1/\psi(t)\geq1/(t \cdot \max_j \gamma^j)\geq 1/t$. Therefore, we derive
\begin{equation}
	\frac{2}{\tilde{\eta}\psi(t)}\mathcal{L}(f^0)\geq\frac{2}{\tilde{\eta}t\max_j\gamma^j}\mathcal{L}(f^0)\geq\frac{2}{\tilde{\eta}t}\mathcal{L}(f^0),
\end{equation}
which means $\min_j\mathbb{S}_{\mathcal{L}}(f^j;{\bm{x}^j}^*)$ has a higher upper bound than $\min_j\mathbb{S}_{\mathcal{L}}(f^j;{\bm{x}^j})$. Note that $\max_j\gamma^j$ is determined by the randomness introduced by sampling and is dependent on $t$. To be specific, $\max_j\gamma^j$ would be close to 0 at the beginning and $\max_j\gamma^j$ approaches 1 as $t$ increases. It means GFT will drop $\mathcal{L}$ faster than RFT at first, which is also demonstrated in Fig.~\ref{tl}.

We see that $t$ is to measure the iteration number of RFT and $\psi(t)$ is to measure that of GFT. Let set $\psi(t)=\tau\leq t$, then we have $t=\psi^{-1}(\tau)$. For RFT, we can derive 
\begin{eqnarray}
    t\geq2\mathcal{L}(f^0)/\left(\tilde{\eta\epsilon}\right).
\end{eqnarray} Therefore, plugging $t=\psi^{-1}(\tau)$ into it and $\psi(\cdot)$ is monotonically increasing, we have $\psi^{-1}(\tau)\geq2\mathcal{L}(f^0)/\left(\tilde{\eta\epsilon}\right)$, that is 
\begin{eqnarray}
    \tau\geq\psi(2\mathcal{L}(f^0)/\left(\tilde{\eta\epsilon}\right)).
\end{eqnarray} $\tau$ measures the iteration number of GFT and $\psi(\frac{2\mathcal{L}(f^0)}{\tilde{\eta}\epsilon})\leq \frac{2\mathcal{L}(f^0)}{\tilde{\eta}\epsilon}$. It means that ITD of GFT is $\mathcal{O}(\psi(\frac{2\mathcal{L}(f^0)}{\tilde{\eta}\epsilon}))\leq\mathcal{O}(2\mathcal{L}(f^0)/\left(\tilde{\eta}\epsilon\right))$.
$\blacksquare$

\section{Detailed Experiments}\label{de}

\begin{figure}[h]
	\vskip -0.1in
	\subfigbottomskip=-6pt
	\centering
	\subfigure[GFT-0.05]{\includegraphics[width=0.49\linewidth]{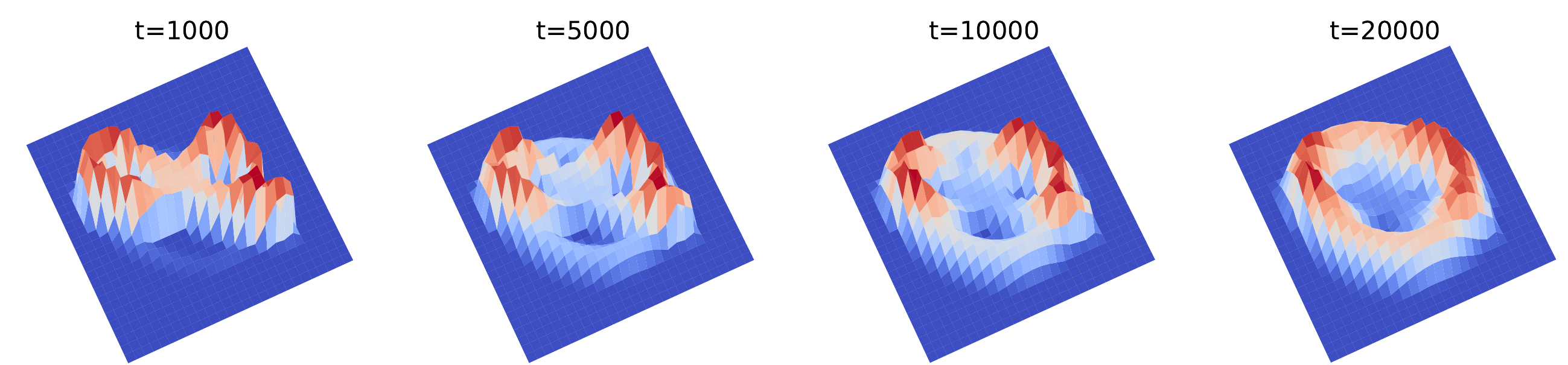}}
	\subfigure[GFT-0.5]{\includegraphics[width=0.49\linewidth]{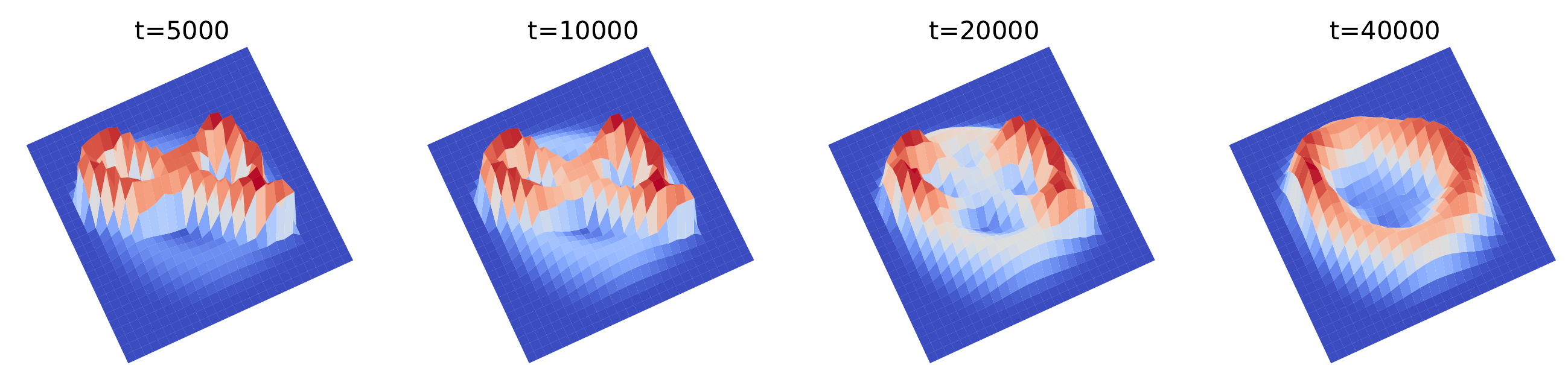}}
	\subfigure[RFT-0.05]{\includegraphics[width=0.49\linewidth]{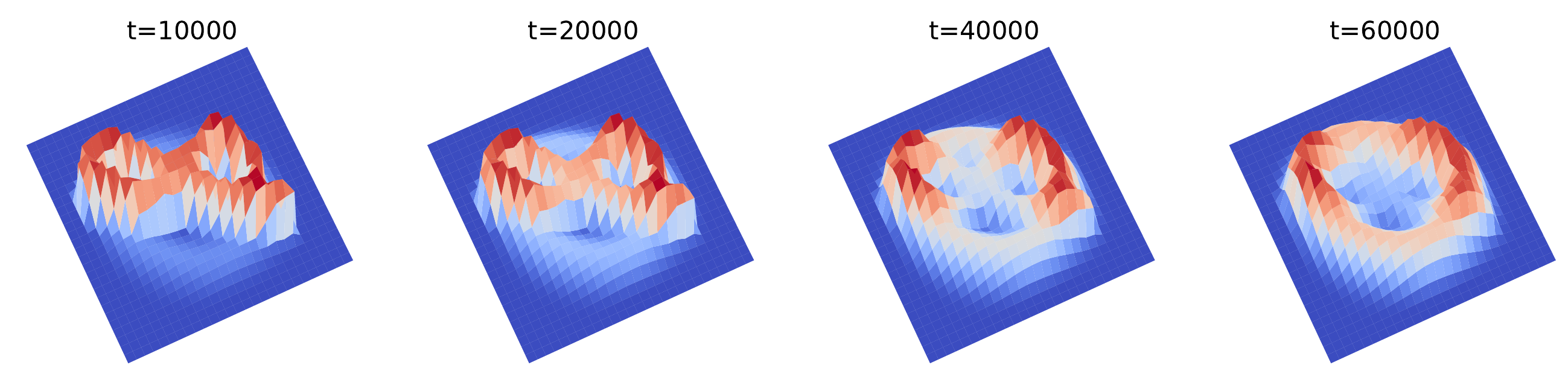}}
	\subfigure[Whole]{\includegraphics[width=0.49\linewidth]{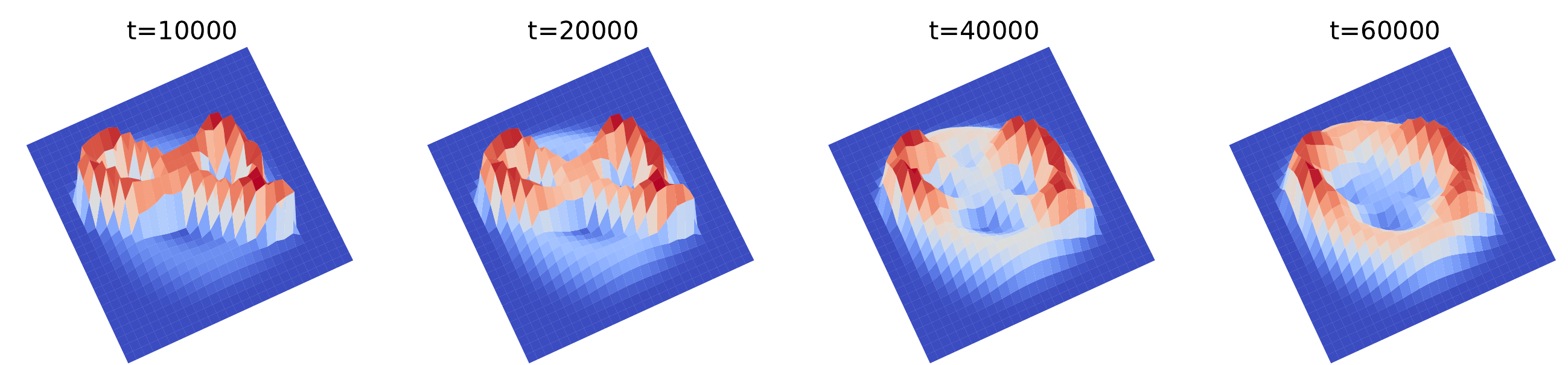}}
	\vskip -2pt
	\caption{Comparing RFT and GFT when nonparametric teaching for correcting 8 towards 0.}
	\label{np80m}
\end{figure}

In computer, operations are discrete. Therefore, we use dense pairwise points $\{(\bm{x}_i,f(\bm{x}_i))\}_{i=1}^n$ to represent a function $f$. For the pool-based teacher (refer to Remark~\ref{pbt}), we use sparse pairwise point to denote $\mathcal{P}$. The pool-based teacher knows $f^*$ but cannot provide some teaching examples out of the pool. For all experiments, we set kernel as the popular and general RBF $K(\bm{x},\bm{x}')=\exp\left(-\left\|\frac{\bm{x}-\bm{x}'}{2}\right\|^2_2\right)$. We specifically take empirical (average) $L_2$ norm defined in Hilbert space to measure the difference between $f$ and $f^*$, 
\begin{equation}
	\mathcal{M}(f,f^*)=\|f-f^*\|_\mathcal{H}=\frac{1}{n}\sqrt{\sum_{i=1}^{n}\left(f(\bm{x}_i)-f^*(\bm{x}_i)\right)^2}.
\end{equation}
Our implementation is based on Intel(R) Core(TM) i7-8750H and NVIDIA GTX 1050 Ti with Max-Q Design.

\textbf{Synthetic 1D Gaussian Mixture.} For this regression problem, we assume the loss function of the learner is square loss $\mathcal{L}=\left(y-f(\bm{x})\right)^2$, and we set it unknown for the teacher. We call the dense pairwise points as pixels which are generated by \texttt{arange(-14, 14, 0.1)}. The learning rate $\eta^t$ is fixed as 0.01. Besides, the teacher will stop if $\mathcal{M}(f^t,f^*)<0.0001$.

\begin{figure*}[t]
	\vskip -0.1in
	\subfigbottomskip=-6pt
	\subfigcapskip=-4pt
	\centering
	\subfigure[Teaching a Parametrized Target Model $y=x+1$.]{\includegraphics[width=\linewidth]{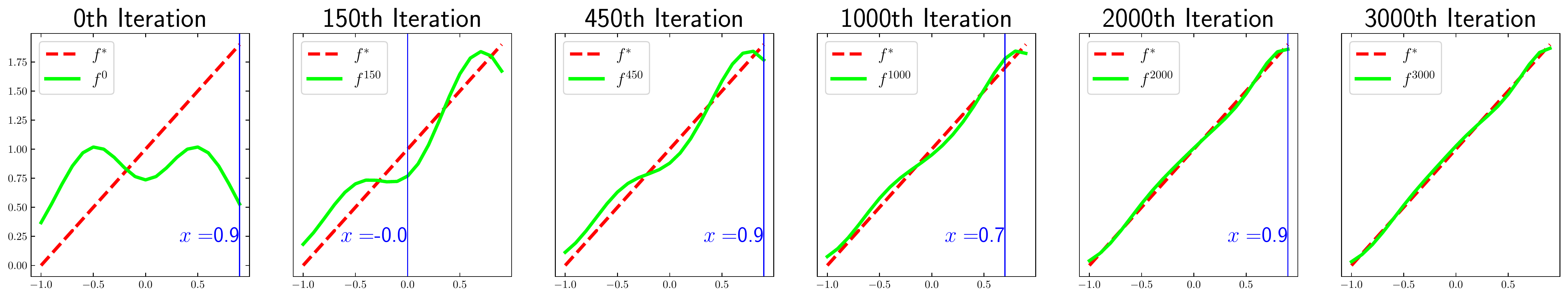}}
	\subfigure[Teaching a Parametrized Target Model $y=-x+1$.]{\includegraphics[width=\linewidth]{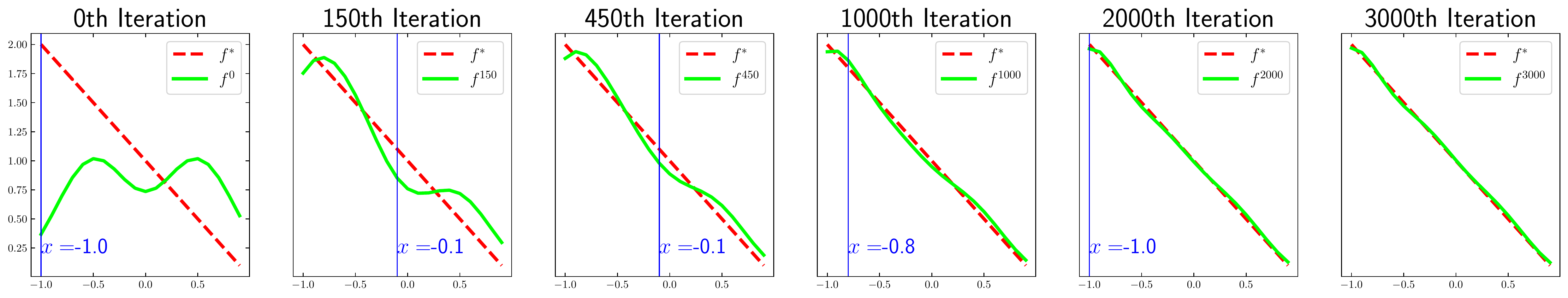}}
	\subfigure[Teaching a Parametrized Target Model $z=x+y-8$.]{\includegraphics[width=\linewidth]{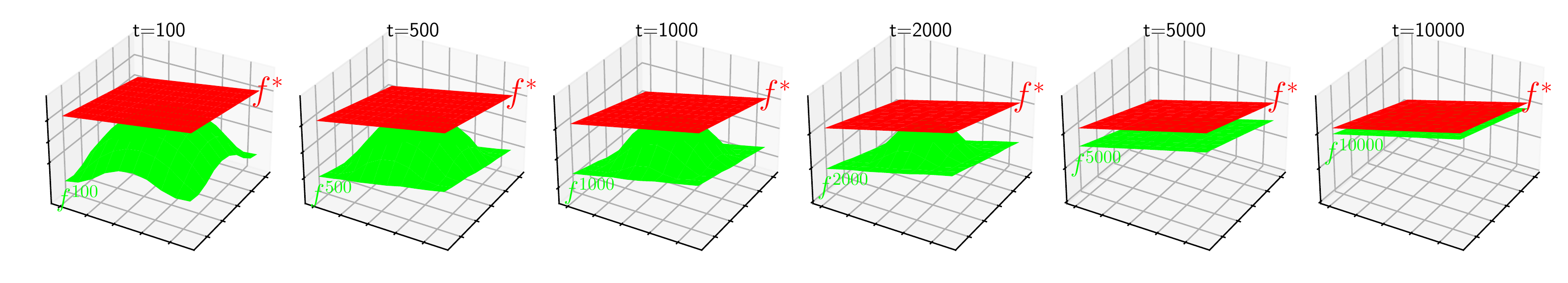}}
	\subfigure[Teaching a Parametrized Target Model $z=x+y-8$.]{\includegraphics[width=\linewidth]{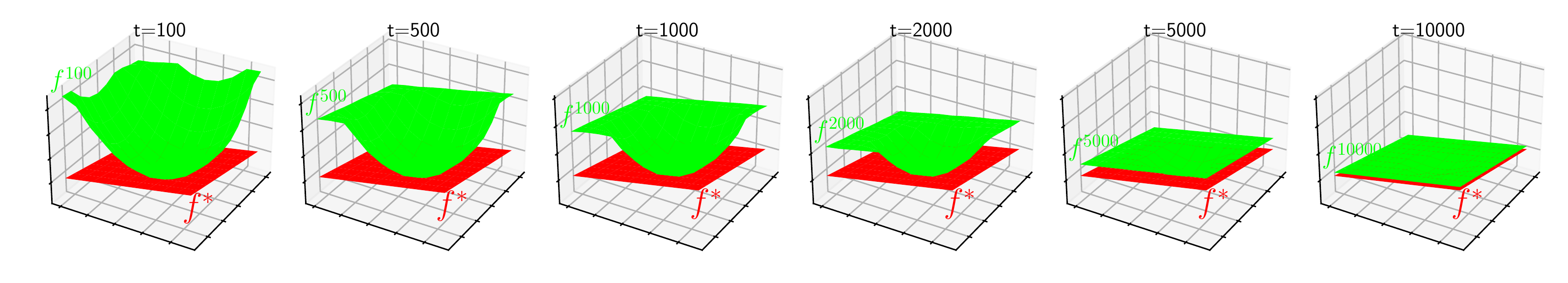}}
	\vskip 0pt
	\caption{GFT for 2D and 3D parameterized target models. (a)-(b): The red dashed lines are $f^*$ and the solid lime lines are $f^t$ at different iteration of GFT. Selected examples are pointed out by blue vertical lines. (c)-(d): The red planes are $f^*$ when $f^t$ is represented by curved lime surfaces. Both 2D and 3D cases show that functional teaching ability of helping the learner converge to $f^*$ even from a bad initial $f^0$ (without overlap with $f^*$), which means that the functional teaching algorithm GFT is well-adapted for parameterized target models.}
	\label{pt}
	\vskip -0.2in
\end{figure*}

\textbf{Synthetic 2D Classification.} For such classification problem, we assume the loss function of the learner is hinge loss $\mathcal{L}=\max\left(0,1-y\cdot f(\bm{x})\right)$ unknown for the teacher. Pixels are generated by \texttt{arange(-1, 1, 0.01)}. The learning rate $\eta^t$ is fixed as 0.001. Besides, the teacher will stop if $\mathcal{M}(f^t,f^*)<0.001$.

\textbf{The digit Correction.} We tend to recover how an infant (the learner) update its opinion about digit 0 when taught. The learning rate $\eta^t$ is fixed as 0.01. $\mathcal{L}=\left(y-f(\bm{x})\right)^2$ is unknown for the teacher. We derive the target $f^*$, optimal 0 via averaging all images of digit 0 in MNIST \cite{lecun1998mnist} (both training and testing sets). We casually pick one digit 8 image as $f^*$. In Figure~\ref{tl}, the loss is $\mathcal{M}(f^t,f^*)$ rather than that of learners $\mathcal{L}$.

In pool-based teaching, the ratio between the pool and the whole sapce can be adjusted, and we set it as 0.8.

In alternative teaching, the alternative of digit 0, character O is selected from EMNIST \cite{cohen2017emnist} via minimizing $\mathcal{M}(f^O,f^*)$, where $f^O$ denotes the character O image. We also scale $f^O$ to match the magnitude of $f^*$ for eliminating influence introduced by the magnitude. the probability of teaching with character O instead of digit 0 can be modified, and we let it as 0.2.

The comparison between RFT and GFT of is presented in Fig.~\ref{np80m}. It shows that for GFT, large k (proportion) would delay the convergence by comparing (a) and (b). Besides, GFT is better than RFT when the hyper parameter k is the same via contrasting (a) and (c). Further, (c)-(d) show that RFT has roughly similar performance as teaching with whole set.

\begin{figure}[th]
	\vskip -0.1in
	\subfigbottomskip=-6pt
	\subfigcapskip=-4pt
	\centering
	\subfigure[GFT-1]{\includegraphics[width=0.49\linewidth]{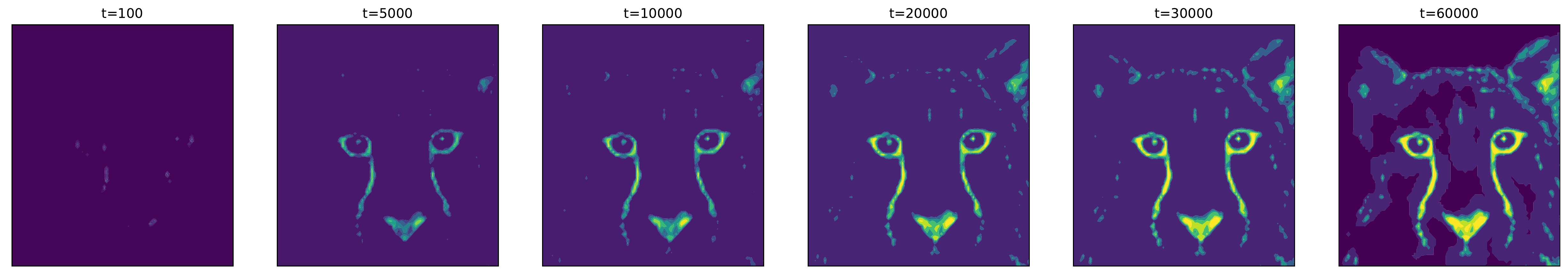}}
	\subfigure[Pool GFT-1]{\includegraphics[width=0.49\linewidth]{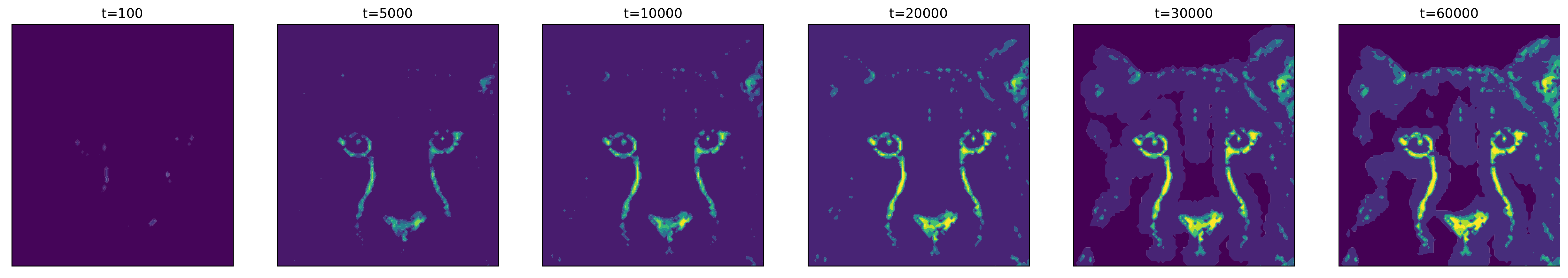}}
	\subfigure[GFT-0.05]{\includegraphics[width=0.49\linewidth]{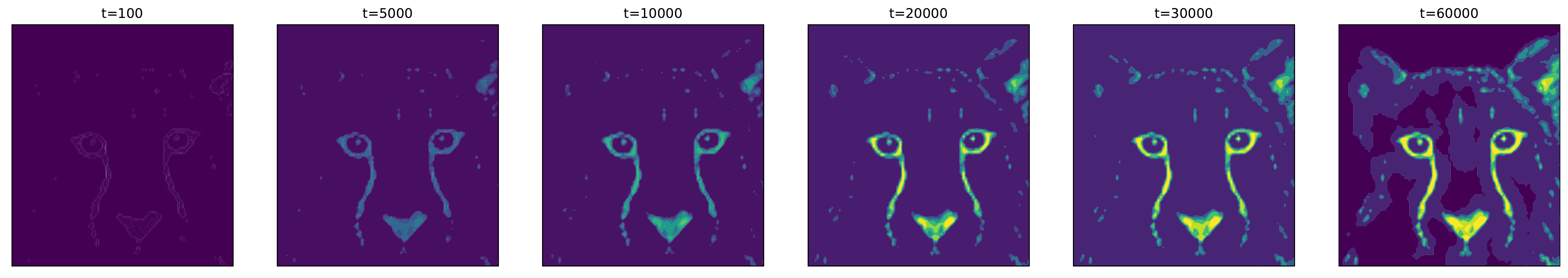}}
	\subfigure[GFT-0.5]{\includegraphics[width=0.49\linewidth]{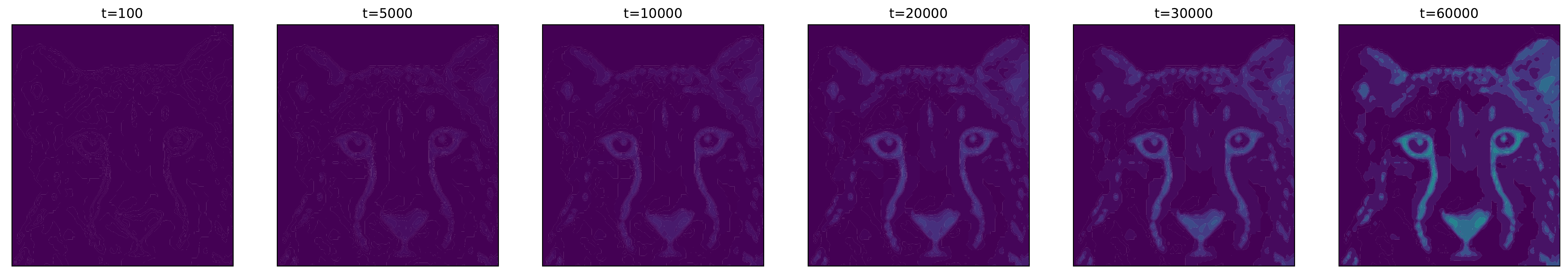}}
	\subfigure[RFT-0.05]{\includegraphics[width=0.49\linewidth]{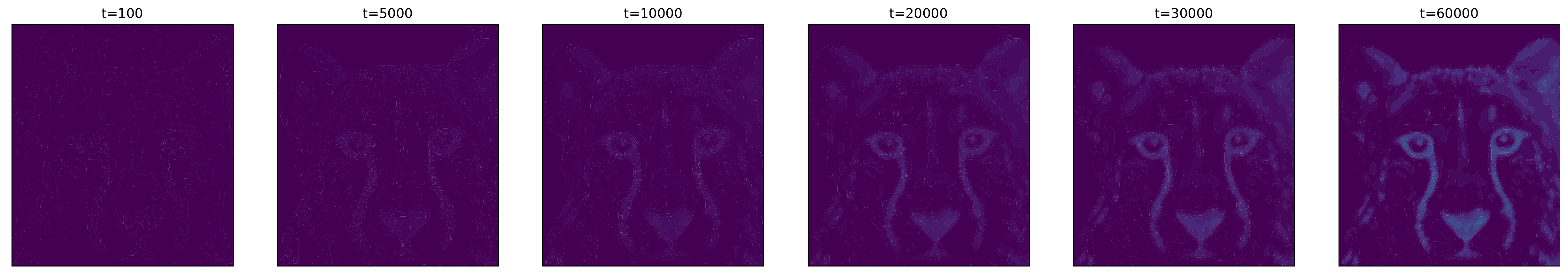}}
	\subfigure[Whole]{\includegraphics[width=0.49\linewidth]{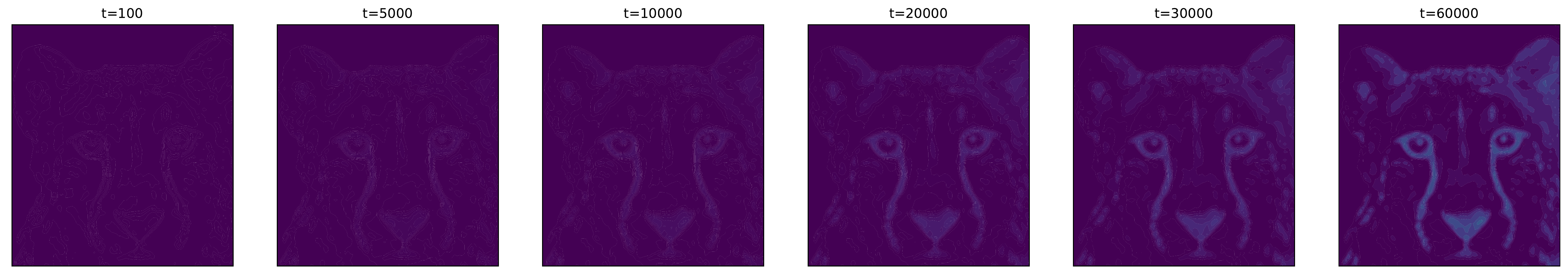}}
	\vskip 0pt
	\caption{Nonparametric teaching for imparting the cheetah. $f$ of GFT become clear significantly faster than RFT. Part of $f$ are not updated for pool-based teaching, so several dark discontinuity points can be found. Moreover, (d)-(f) presents a smooth performance, which indicates that pack teaching can effectively smooth the gradient for the convergence to the target model.}
	\label{nchecountour}
	\vskip -0.1in
\end{figure}

\textbf{The cheetah impartation.} The learning rate $\eta^t$ is fixed as 0.01. $\mathcal{L}=\left(y-f(\bm{x})\right)^2$ is unknown for the teacher. We derive this cheetah figure from \citealp{shen2020sinkhorn} who use pickles to sketch. Differently, we regard a figure as a smooth function and impart it to learners via functional teaching algorithms RFT and GFT. Fig.~\ref{nchecountour} is the contour version of Fig.~\ref{nche}.

\section{Experiment Extensions} \label{ee}

\textbf{Teaching parametric target models from nonparametric initialization with GFT.}
We further test parametric adaptation of GFT. Specifically, we let the target model is identified by the parameter but remain teaching function directly instead of its parameter to see the performance of GFT. Here, we assume the loss function of the learner is square loss $\mathcal{L}=\left(y-f(\bm{x})\right)^2$. In two 2D cases, we set $f^*(x)=x+1$ and $f^*(x)=-x+1$, respectively when both $f^0(x)=\exp(-\left(\frac{x - 0.5}{0.5} \right)^2) + \exp(-\left(\frac{x + 0.5}{0.5} \right)^2)$. The learning rate $\eta=0.01$ and pixels are generated by \texttt{arange(-1, 1, 0.1)}. Besides, the teacher will stop if $\mathcal{M}(f^t,f^*)<0.0001$. We see that in Fig.~\ref{pt} (a)-(b) even the target model is a straight line while the initial one is a curve, GFT also can straighten $f^t$ and cover $f^*$ approximately. In two 3D cases, we let both $f^*(x_1,x_2)=x_1+x_2-8$, and let $f^0=-\left(x_1-5\right)^2-\left(x_2-5\right)^2$ and $f^0=\left(x_1-5\right)^2+\left(x_2-5\right)^2$, respectively. The learning rate $\eta=0.01$ and $x_1,\,x_2$ pixels are generated by \texttt{arange(0, 10, 1)}. We observe that in Fig.~\ref{pt} (c)-(d) when the target model is identified by the vector $(1,1,-8)^T$, GFT is also able to teach curved surfaces towards this plane. To summarize, the functional teaching algorithm GFT is well-adapted for parameterized target models and GFT could teach the target function beyond its parameter.

\begin{figure*}[t]
	\subfigbottomskip=-6pt
	\subfigcapskip=-4pt
	\centering
	\includegraphics[width=\linewidth]{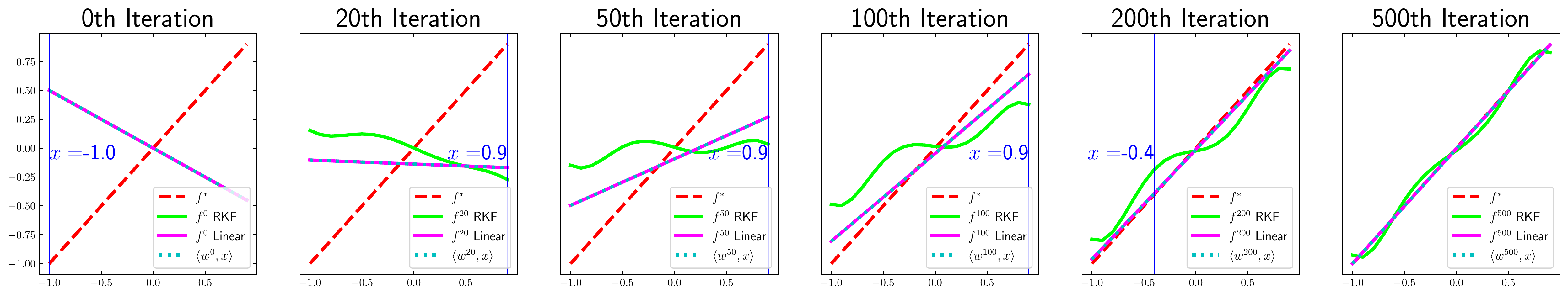}
	\vskip -3pt
	\caption{Contrast nonparametric teaching with the RBF and Linar kernels against parametric teaching. For fairness, the fed teaching examples are all from GFT. The red dashed lines are $f^*$. Nonparametric teaching: the solid lime lines are $f^t$ with RBF kernels and the solid magenta lines are $f^t$ with Linear kernels. Parametric teaching: The dotted lines are $f^t = \langle w^t,\bm{x} \rangle$. $f^t$ in all settings converge fast. Interestingly, nonparametric teaching with Linear kernel has same performance as parametric teaching in each round. This is reasonable because the contribution of functional gradient under the Linear kernel is just updating $w^t$ from the parameterized viewpoint. It concludes that nonparametric teaching is more general than parametric teaching.}
	\label{lk}
\end{figure*}

\textbf{The comparison between nonparametric and parametric teaching under parameterized initialization.} We set $\eta^t=0.01$, $\mathcal{L}=\left(y-f(\bm{x})\right)^2$ and $f^*= \langle w^*,\bm{x}\rangle = \langle (1,1)^T,(x,1)^T\rangle =x+1$ and $f^0= \langle w^0,\bm{x}\rangle = \langle (-0.5,0.5)^T,(x,1)^T\rangle =-0.5x+0.5$. For nonparametric teaching, except for RBF kernel defined before, we introduce a Linear kernel $K(\bm{x},\bm{x}')=\langle\bm{x},\bm{x}' \rangle+1$. In each iteration, we let GFT-1 select a teaching example and learners evolve $f^t$ based on RBF and Linear kernels respectively. For parametric teaching, we let learners use parameter gradient descent:
\begin{equation}
	w^{t+1}\gets w^t-\eta^t\mathcal{G}(\mathcal{L};w^t;(\bm{x}^t,y^t)).
\end{equation}
For fairness, the provided teaching examples are the same as that of nonparametric teaching derived by GFT-1. We observe that $f^t$ in both nonparametric and parametric teaching converge fast. Interestingly shown in Fig.~\ref{lk}, we find that nonparametric teaching with Linear kernel has same results as parametric teaching in every iteration. This is under expectation since the influence of functional gradient under the Linear kernel in each iteration is just modifying $w^t$ from the parameterized viewpoint. \textit{This means parametric teaching could be viewed as a particular case of nonparametric teaching when kernel is a Linear one} $K(\bm{x},\bm{x}')=\langle\bm{x},\bm{x}' \rangle+c$, \textit{where $c$ is a constant}.

\begin{figure}[t]
	\centering
	\includegraphics[width =\linewidth]{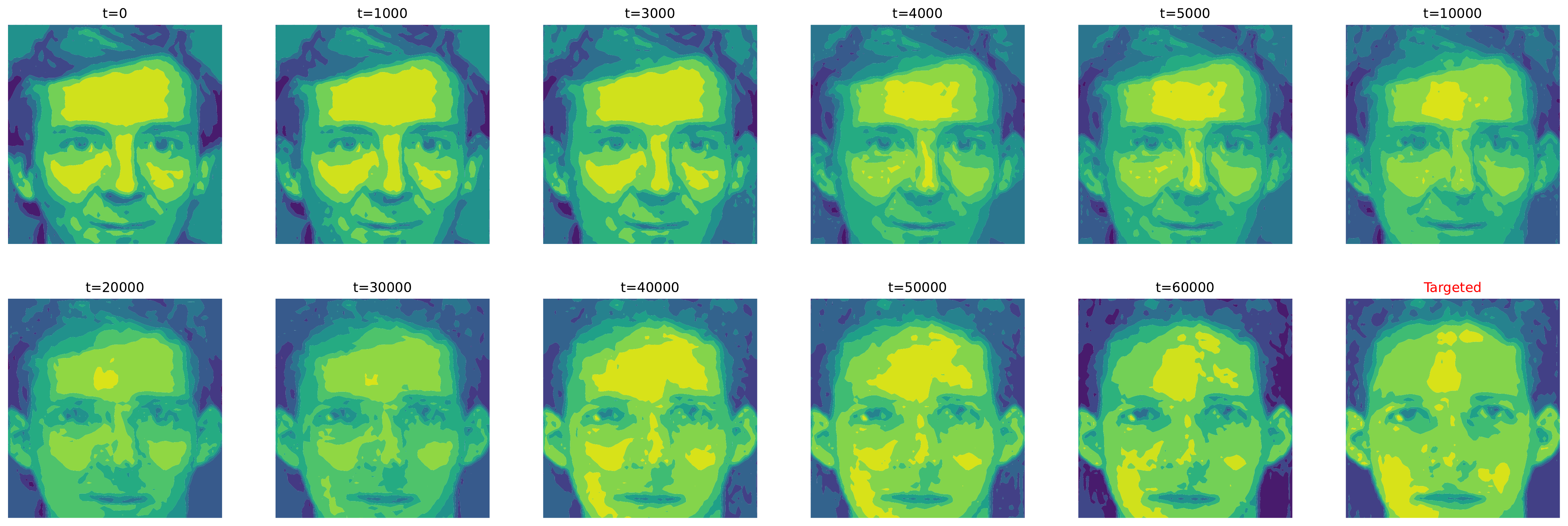}
	\vskip -4pt
	\caption{GFT for facial teaching. The left top one is the initial photograph and the right bottom one is the target. Viewing the facial photograph as the function, GFT works well. }
	\label{hf}
	\vskip -0.2in
\end{figure}

\textbf{The sketch for missing person report.} Consider a practical and interesting scenario that associates wish to file a missing person report at a police station without a photograph. The police considered as the learner would randomly provide a initial photograph, then associates (the teacher) can update the initial photograph based on their impressions in mind, which is precisely a teaching process.

GFT is also applicable in above task as a smooth solution. Smooth means $f^t$ is modified gradually instead of replacing. To handle above nonparametric teaching problems, one can view the human face in the photograph as a general function, and GFT would modify the initial one, \ie, random initialization from police, towards the targeted one (the image of the missing person in associates' minds), which is shown in Fig.~\ref{hf}. Specifically, we pick two facial figures form the ORL database (http://www.cam-orl.co.uk), then we set one as initialization and the other as target. The learning rate $\eta^t$ is fixed as 0.05. $\mathcal{L}=\left(y-f(\bm{x})\right)^2$ is unknown for the teacher. We see that even for the complicated facial figure, our GFT presents expected performance.

\end{document}